\documentclass[letterpaper]{article}
\usepackage{uai2020}
\usepackage[margin=1in]{geometry}

\usepackage{times}
\usepackage{natbib}
\usepackage{hyperref}
\usepackage{graphicx}
\usepackage{amsmath}
\usepackage{amsfonts}
\usepackage{amsthm}
\usepackage{multicol}
\usepackage{multirow}
\usepackage{appendix}
\usepackage{color}

\newtheorem{definition}{Definition}
\newtheorem{remark}{Remark}

\setcitestyle{round}

\title{Towards Threshold Invariant Fair Classification}

\author{ {\bf Mingliang Chen and Min Wu} \\
Electrical and Computer Engineering Department and Institute for Advanced Computer Studies\\
University of Maryland, College Park, MD, USA \\
\{mchen126, minwu\}@umd.edu
}

\begin{document}

\maketitle

\begin{abstract}
Effective machine learning models can automatically learn useful information from a large quantity of data and provide decisions in a high accuracy. These models may, however, lead to unfair predictions in certain sense among the population groups of interest, where the grouping is based on such sensitive attributes as race and gender. Various fairness definitions, such as demographic parity and equalized odds, were proposed in prior art to ensure that decisions guided by the machine learning models are equitable. Unfortunately, the ``fair'' model trained with these fairness definitions is threshold sensitive, i.e., the condition of fairness may no longer hold true when tuning the decision threshold. This paper introduces the notion of threshold invariant fairness, which enforces equitable performances across different groups independent of the decision threshold. To achieve this goal, this paper proposes to equalize the risk distributions among the groups via two approximation methods. Experimental results demonstrate that the proposed methodology is effective to alleviate the threshold sensitivity in machine learning models designed to achieve fairness.

\end{abstract}

\section{INTRODUCTION}
Machine learning is being used across a wide variety of practical tasks in recent years, influencing many aspects of our daily life. Many companies and government departments have deployed machine learning based decision-making systems to facilitate decision making in business operations. The U.S. courts use a software known as Correctional Offender Management Profiling for Alternative Sanctions (COMPAS), to measure the risk for a defendant to recommit another crime, which helps judges make parole decisions. Banks turn to machine learning models to evaluate the prospective customers' credits and to help decide the approval of loan applications. Like humans, machine learning algorithms can suffer from bias that render their decisions ``unfair''~\citep{barocas2016big,mehrabi2019survey}. A number of investigations have shown that the existing machine learning systems have fairness issues in certain sense. For instance, the works by~\citep{angwin2016machine,larson2016we} studied the COMPAS software and found a bias against African-Americans: it is more likely to assign a higher risk score to an African-American defendants than to a Caucasian with a similar profile.

In the context of decision making, fairness typically means the equivalent (non-discriminatory) decision performances across different groups of people based on their inherent or acquired characteristics. From this point of view, a number of recent studies introduced {\it demographic parity} (DP)~\citep{dwork2012fairness,feldman2015certifying} and {\it equalized odds} (EO)~\citep{hardt2016equality} to characterize and evaluate the fairness level in the machine learning models. Since the goal of these definitions is to equalize certain probability statistics computed from the confusion matrix of classification, this type of definitions were named as {\it classification parity} in \citep{corbett2018measure}. To take the COMPAS case as an example, DP means that parole granting rates should be equal across the race groups; and EO enforces that among defendants who would or would not have gone on to commit a violent crime if released, the rates are equal across the race groups.

The fairness algorithms in the literature mainly fall into three categories: 1) Preprocessing: the studies by~\citep{zemel2013learning,calmon2017optimized} map the data to a fair representation in a latent space satisfying the defined fairness or independent of the protected attribute. Ideally, removing correlated features and finding independent representation to the protected attribute may lead to undesirable decision outcomes. Imagine a company scores randomly on the applicants over two groups. Even though the scoring function is independentof the groups, it is likely that a considerable number of unqualified applicants are selected at last. 2) Constraints on training: the works by~\citep{zafar2017eo,zafar2017dp} constrain the classifier on DP and EO, respectively, during the training. Fair classification can also be reduced to a set of sub-problems~\citep{agarwal2018reductions}. The tradeoff is required between accuracy and fairness constraints. 3) Postprocessing: the work by~\citep{hardt2016equality} searches proper thresholds on scores over different groups. The studies by~\citep{crowson2016assessing,pleiss2017fairness} calibrate on the prediction score such that the probability of positive label is equal to the prediction score. The postprocessing requires access to protected attributes in the test phase.

Classification parity can help machine learning models achieve equitable decision performances across the groups from a probability point of view, such as positive proportion of decisions, true positive rate, false positive rate, and other statistics alike. However, such parity of the group-wise performances is generally not retained, when we tune the decision threshold. Trained under the constraint of classification parity, the classifiers only ``delicately'' achieve fairness requirement in the default decision thresholds, and are sensitive and vulnerable to the change of decision thresholds. We will illustrate the limitation of classification parity in Section~\ref{sec:limit}.

To alleviate the limitation of classification parity, we introduce a new notion of {\it threshold invariant fairness} in this paper. Unlike classification parity, threshold invariant fairness enforces a stronger condition on the classifier such that it has a consistent fairness level of classification results against the change of decision threshold.

To develop our proposed notion, we find that it is sufficient to design the scoring function of the classifier to equalize the distributions of the risk scores over all groups. From classification parity to threshold invariant fairness, the equalization focus shifts from statistical attributes at a given decision rule to the probability distribution of the risk scores, suggesting that our proposed notion imposes a stronger constraint than that by the classification parity. In this paper, we propose two approximation approaches to equalize the distributions of risk scores over all groups and incorporate them into a variety of differentiable classifiers, such as logistic regression and support vector machine. The problem can be solved by a gradient-based optimizer. Our new methods have two advantages compared with the prior art: 1) we do not require the protected attributes in the test phase; 2) we can tune the threshold to modify the positive rate of the classifier predictions and maintain the fairness as much as possible.

\section{BACKGROUND}
\subsection{CLASSIFICATION PARITY}
\label{sec:def}

Classification parity is an approach to alleviate the machine bias by equalizing performance measure of classification across the groups with the protected attributes to achieve a group-wise fair classification. As early reviewed, {\it demographic parity}~\citep{dwork2012fairness,feldman2015certifying} and {\it Equalized Odds}~\citep{hardt2016equality} are two well-known fairness definitions in classification parity. We assume that one input sample with observed $m$-dimensional features $X\in\mathbb{R}^m$ has a protected attribute $A\in\{0,1\}$, and a groud truth label $Y\in\{0,1\}$; and $\hat{Y}\in\{0,1\}$ is the classifier's prediction. In a simplified recidivism example, $A$ may represent the defendant's race as African-American versus Caucasian, $Y$ represents whether the defendant recommits another crime after being released. The definition of two existing classification parities are described as follows.

\begin{definition} (Demographic Parity). A predictor $\hat{Y}$ satisfies demographic parity if
\begin{equation}
    P(\hat{Y}=1|A=0)=P(\hat{Y}=1|A=1).
\end{equation}
\end{definition}
\begin{definition} (Equalized Odds). A predictor $\hat{Y}$ achieves equalized odds if the true positive rates and false positive rates across the groups are equal, respectively, i.e.,
\begin{equation}
\left\{\begin{matrix}
    P(\hat{Y}=1|A=0,Y=0)=P(\hat{Y}=1|A=1,Y=0),\\
    P(\hat{Y}=1|A=0,Y=1)=P(\hat{Y}=1|A=1,Y=1).
\end{matrix}\right.
\end{equation}
\end{definition}
We see that DP enforces that the selection rates are equal across all the protected attributes; EO enforces that the prediction accuracy is equally high across all the protected attributes. Based on the above parity definition, many papers have proposed algorithms to achieve classification parity via pre-processing~\citep{kamiran2012data,zemel2013learning}, post-processing~\citep{hardt2016equality}, and regularization techniques~\citep{dwork2012fairness,feldman2015certifying,zafar2017eo,zafar2017dp,corbett2017algorithmic,agarwal2018reductions}.

\subsection{LIMITATION OF THE PRIOR ART}
\label{sec:limit}

\begin{figure}[t]
\centering
\includegraphics[width=0.8\linewidth]{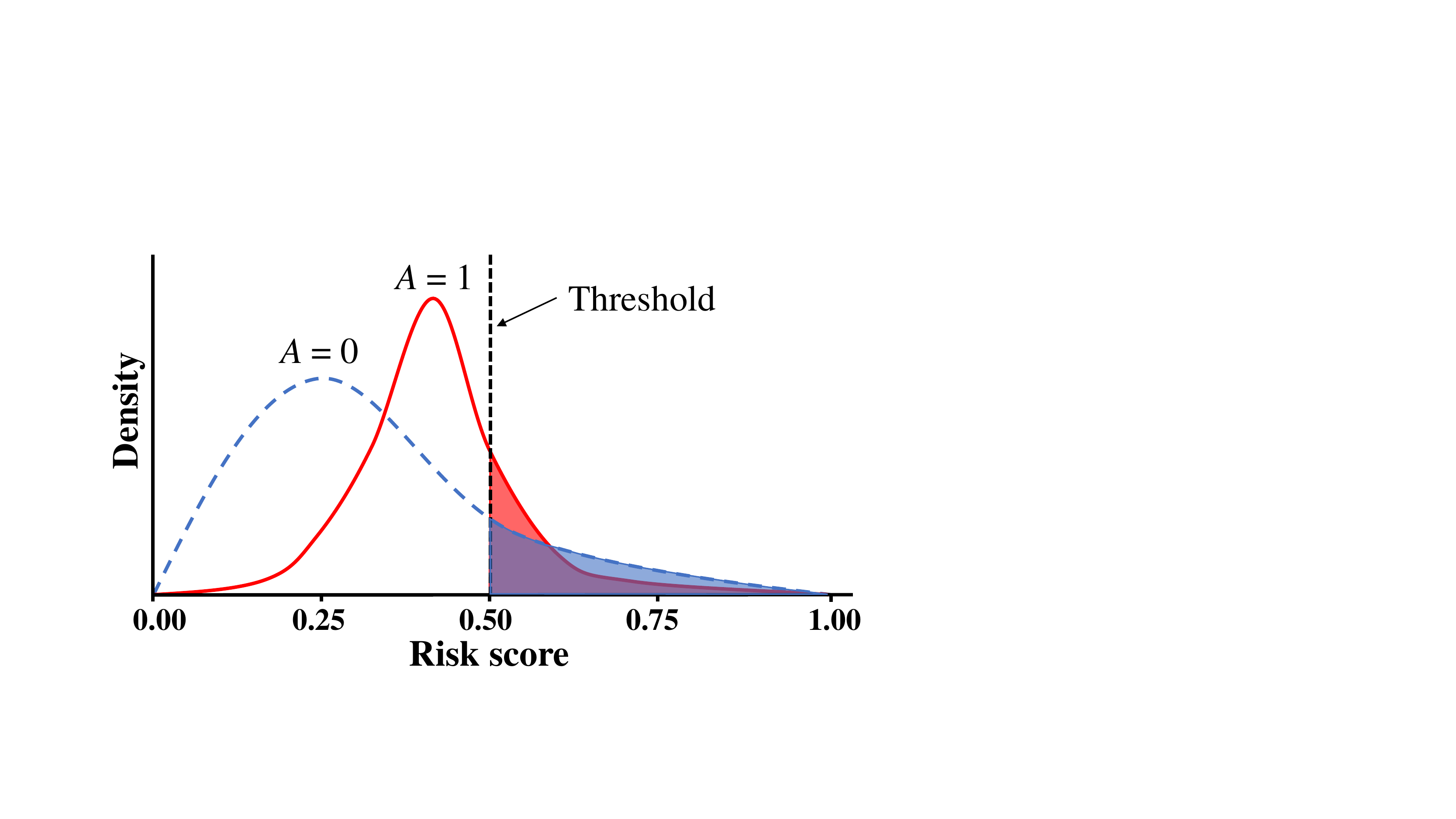}
\caption{Hypothetical risk distributions of two groups (e.g., $A=0$ and $A=1$). When classification parity is satisfied, i.e., the areas of the highlighted regions are equal, the risk distributions can still differ from each other. Classification parity does not achieve equitable decision performance in terms of risk distribution.}
\label{fig:risk}
\end{figure}

As pointed out by \citep{corbett2018measure}, classification parity is a problematic measure of fairness. We consider a risk score $s(X) = P(\hat{Y}=1|X)$, describing the probability that an input sample $X$ will be predicted positive. Given a group $\mathcal{D}_a$ with protected attribute $A=a$, we can compute the distribution of $s(X)$ over the group $\mathcal{D}_a$, which we refer to as risk distribution $r(\mathcal{D}_a)$. Figure~\ref{fig:risk} shows hypothetical risk distributions for two groups with different protected attributes. In the model inference stage, an input sample is classified by comparing its risk score with a decision threshold. The proportion of the positive predictions in one group is the fraction of the risk distribution that is to the right of the decision threshold, highlighted in Figure~\ref{fig:risk}. 

According to the definitions reviewed in Section~\ref{sec:def}, to achieve classification parity is to equalize the positive (or negative) proportions among different groups. However, when the classifier achieves classification parity at a specific decision threshold, i.e., the areas of the highlighted regions are equal, the risk distributions can still differ from each other. This suggests that the classification parity can be easily destroyed by changing the decision threshold. The ``fair'' machine learning model under classification parity does not show equitable performances among different groups in terms of the risk distribution. The goal of this paper is to propose a stronger condition on fairness to alleviate the above issue, and to develop systematic method by designing the scoring function during the classifier training to achieve it. 
\section{THRESHOLD INVARIANT FAIRNESS}
\label{sec:fairness}
\subsection{NOTATIONS}
Recall that an input sample with $m$-D features $X\in\mathbb{R}^m$ has protected attribute $A\in\{0,1\}$, its corresponding ground truth label $Y\in\{0,1\}$, and the classifier prediction is $\hat{Y}\in\{0,1\}$. Define the scoring function in classifier $g:\mathbb{R}^m\mapsto\mathbb{R}$, mapping the $m$-D features to a scalar, the raw score. A {\it sigmoid} activation $\sigma(\cdot)$ is applied to confine the raw score in the range of $[0,1]$, which can be interpreted as the risk score defined in Section~\ref{sec:limit}, i.e., $s(X)=\sigma(g(X))$. Given the threshold $t\in[0,1]$, the prediction of the sample $X$ is determined as 
\begin{equation}
\hat{Y}=\left\{\begin{matrix}
0, & s(X)\leq t,\\ 
1, & s(X)>t.
\end{matrix}\right.
\end{equation}

\subsection{FAIRNESS DEFINITION}
We define threshold invariant fairness, a stronger condition on classification parity, in the context of DP and EO.

\begin{definition} (Threshold Invariant Fairness). Threshold Invariant Demographic Parity (TIDP) or Threshold Invariant Equalized Odds (TIEO) is achieved when DP or EO is satisfied, respectively, independent of the decision threshold $t$.
\end{definition}

We use the Calder-Verwer (CV) score~\citep{calders2010three} to measure classification parity. For simplicity, we denote that $P_a(\hat{Y}):=P(\hat{Y}|A=a)$ and $P_{a,y}(\hat{Y}):=P(\hat{Y}|A=a,Y=y)$. The CV scores for DP and EO are defined as
\begin{equation}
\begin{aligned}
    \Delta \textup{DP}=&\big |P_0(\hat{Y}=1)-P_1(\hat{Y}=1)\big |,\\
    \Delta \textup{EO}=&\frac{1}{2}(\big |P_{0,0}(\hat{Y}=1)-P_{1,0}(\hat{Y}=1)\big |\\
    &+\big |P_{0,1}(\hat{Y}=1)-P_{1,1}(\hat{Y}=1)\big|).
\end{aligned}
\end{equation}
Note that the smaller the CV score is, the better the classifier achieves a fair classification. In the most ideal cases, the classification parity, DP or EO, is satisfied when $\Delta \textup{DP}$ or $\Delta \textup{EO}$ is zero.

In the following, we investigate the relationship between threshold invariant fairness and the risk distribution. Denote $f_{a}(s):=f(s|A=a)$ and $f_{a,y}(s):=f(s|A=a,Y=y)$ as the risk distributions over the group with the protected attribute $A$ and label $Y$. Recall that the support of the risk score $s$ is $[0,1]$ and the decision threshold $t$ ranges within $[0,1]$. For the DP constraint, we have
\begin{equation}
\begin{array}{l}
P_a(\hat{Y}=1)=P_a(s(X)>t)=\int_{t}^{1}f_{a}(s)\textup{d}s,\\
P_a(\hat{Y}=0)=P_a(s(X)\leq t)=\int_{0}^{t}f_{a}(s)\textup{d}s.
\end{array}
\end{equation}
Then,
\begin{equation}
\begin{aligned}
\Delta \textup{DP}&=\big |P_0(\hat{Y}=1)-P_1(\hat{Y}=1)\big |\\
&=\big|\int_{t}^{1} \big(f_{0}(s)-f_{1}(s)\big)\textup{d}s\big |\\
&\leq\int_{t}^{1} \big|f_{0}(s)-f_{1}(s)\big |\textup{d}s\leq\epsilon_{DP}(1-t).
\end{aligned}
\label{eqn:dp1}
\end{equation}
where we define the upper bound of the difference between two risk distributions $\big|f_{0}(s)-f_{1}(s)\big|\leq\epsilon_{DP},\forall s\in[0,1]$. Since $P_a(\hat{Y}=1)-P_a(\hat{Y}=0)=1,\forall a\in\{0,1\}$, we have
\begin{equation}
\begin{aligned}
\Delta \textup{DP}&=\big |P_1(\hat{Y}=0)-P_0(\hat{Y}=0)\big |\\
&\leq\int_{0}^{t} \big|f_{1}(s)-f_{0}(s)\big |\textup{d}s
\leq\epsilon_{DP}t.
\end{aligned}
\label{eqn:dp2}
\end{equation}
Combining (\ref{eqn:dp1}) and (\ref{eqn:dp2}), we have
\begin{equation}
\Delta \textup{DP}\leq\epsilon_{DP}\cdot\min\{t,1-t\}\leq\frac{1}{2}\epsilon_{DP}.
\end{equation}
\begin{remark} (The sufficient condition of TIDP). When $\epsilon_{DP}$ reaches 0, DP is satisfied regardless of the decision threshold. Hence, equalizing the risk distributions between the positive and negative protected attributes, i.e., $f_0(s)=f_1(s)$, is the sufficient condition of TIDP.
\end{remark}
Similarly, for EO constraint, we have
\begin{equation}
\Delta \textup{EO}\leq\frac{1}{2}(\epsilon_0+\epsilon_1)\cdot\min\{t,1-t\}\leq\frac{1}{4}(\epsilon_0+\epsilon_1).
\label{eqn:delta_eo}
\end{equation}
where we define the upper bounds $\big|f_{0,0}(s)-f_{1,0}(s)\big |\leq\epsilon_0$ and $\big|f_{0,1}(s)-f_{1,1}(s)\big |\leq\epsilon_1, \forall s\in[0,1]$. The derivation is given in Appendix A.
\begin{remark} (The sufficient condition of TIEO). When $\epsilon_0$ and $\epsilon_1$ reach 0, EO is satisfied regardless of the decision threshold. Hence, equalizing the risk distributions of the positive/negative samples between positive and negative protected feature attributes, i.e., $f_{0,y}(s)=f_{1,y}(s),~\forall y\in\{0,1\}$, respectively, is the sufficient condition of TIEO.
\end{remark}
Instead of equalizing statistical attributes, threshold invariant fairness requires equalization of risk distributions, which is a stricter condition than the family of classification parity.

\section{PROPOSED FAIR CLASSIFIER}
\label{sec:model}

So far, we know that equalizing risk distributions across different groups is a sufficient condition to achieve threshold-invariant fairness. One straightforward post-processing approach is to find a mapping function to equalize the histogram of risk scores over one group to another group, a strategy similar to histogram matching in image processing~\citep{gonzalez2004digital}. Here, we focus on holistic systematic approaches to equalize risk distributions during the classifier training by formulating an appropriate fairness regularization. We start with the formulation of the proposed fairness in Section~\ref{sec:equa}, and then show how to incorporate the regularizer into the classifier design in Section~\ref{sec:formulation}.

\subsection{EQUALIZATION OF RISK DISTRBUTION}
\label{sec:equa}
In this subsection, we present two approaches to design the regularizer between two groups $\mathcal{D}_0$ and $\mathcal{D}_1$ in the whole sample set $\mathcal{D}$.
\paragraph{Approach 1: Histogram Approximation (HA)} ~

Define $C=\{c_j\}_{j=1}^{N}$ as the bin centers of $N$-bin histogram with equal interval and the bin width $\delta=c_j-c_{j-1}$. The count of the samples from the group $\mathcal{D}_i$ in the bin $(c-\frac{\delta}{2},c+\frac{\delta}{2})$ is expressed as
\vspace{-1mm}
\begin{equation}
    n_c=\sum_{X\in\mathcal{D}_i}\Pi_c(s(X)),
\vspace{-1mm}
\end{equation}
where $\Pi_c(\cdot)$ is a rectangular function:
\begin{equation}
    \Pi_c(x)=
    \left\{\begin{matrix}
1,&x\in(c-\frac{\delta}{2},c+\frac{\delta}{2}),\\ 
0,&\textup{otherwise}.
\end{matrix}\right.
\end{equation}
Since $\Pi_c(\cdot)$ is not differentiable, we approximate it with a Gaussian kernel,
\begin{equation}
\footnotesize
    n_c=\sum_{X\in\mathcal{D}_i}G_c(s(X))=\sum_{X\in\mathcal{D}_i}\exp(-\frac{(s(X)-c)^2}{2\sigma_c^2}).
\label{eq:gauss}
\end{equation}
Hence, the histogram of the risk score in $\mathcal{D}_i$ can be expressed as
\begin{equation}
    h(\mathcal{D}_i)=normalize\big(\left[n_{c_1},n_{c_2},...,n_{c_N}\right]\big),
\end{equation}
where $normalize(\cdot)$ scales the histogram such that the histogram entries sum to one. Note that $h(\mathcal{D}_i)$ is an $N$-D vector since the histogram has $N$ bins. The HA method formulates a differentiable histogram via a Gaussian kernel to construct the risk distribution.

We use the symmetric combination of Kullback--Leibler (KL) divergence to evaluate the distance of the risk distributions between two groups $\mathcal{D}_0$ and $\mathcal{D}_1$.
\begin{equation}
\begin{aligned}
&d\big(r(\mathcal{D}_0),r(\mathcal{D}_1)\big)\\
=&D_{KL}\big(h(\mathcal{D}_0)||h(\mathcal{D}_1)\big)+D_{KL}\big(h(\mathcal{D}_1)||h(\mathcal{D}_0)\big)\\
=&\sum_{i=1}^{N}\big(h_{(i)}(\mathcal{D}_0)-h_{(i)}(\mathcal{D}_1)\big)\log\frac{h_{(i)}(\mathcal{D}_0)}{h_{(i)}(\mathcal{D}_1)}.
\end{aligned}
\label{eq:dist1}
\end{equation}
where subscript $(i)$ denotes the $i$-th entry of the histogram. It can be seen that the distance $d$ is non-negative and approaches zero when two risk distributions are equal, i.e., $r(\mathcal{D}_0)=r(\mathcal{D}_1)$.

\paragraph{Approach 2: Gaussian Assumption (GA)} ~

Since the risk scores are the sigmoid of the raw scores, i.e., $s(X)=\sigma(g(X))$, equalizing the distribution of raw scores would lead to the equalization of the risk distribution. Given two groups $\mathcal{D}_0$ and $\mathcal{D}_1$, we assume that the distributions of raw scores obey the same Gaussian distribution over $\mathcal{D}_0$ and $\mathcal{D}_1$, and the mean and variance are sufficient to characterize the distributions. We estimate means and variances of the distributions of raw score over $\mathcal{D}_0$ and $\mathcal{D}_1$ using maximum likelihood estimate:
\begin{equation}
\mathcal{D}_i:\left\{\begin{array}{l}
    \hat{\mu}_i=\frac{1}{|\mathcal{D}_i|}\sum_{X\in\mathcal{D}_i}g(X)\\
    \vspace{-7pt}
    \\\hat{\sigma}_i^2=\frac{1}{|\mathcal{D}_i|}\sum_{X\in\mathcal{D}_i}\big(g(X)-\hat{\mu}_i\big)^2
    \end{array}\right.,~i\in\{0,1\}.
\end{equation}
Recall that KL divergence of two 1-D Gaussian distributions $\mathcal{N}_0(\mu_0,\sigma_0^2)$ and $\mathcal{N}_1(\mu_1,\sigma_1^2)$ is
\begin{equation}
\footnotesize
D_{KL}\big(\mathcal{N}_0||\mathcal{N}_1\big)=\frac{1}{2}\big(\log\frac{\sigma_1^2}{\sigma_0^2}+\frac{(\mu_0-\mu_1)^2+\sigma_0^2}{\sigma_1^2}-1\big).
\label{eq:gauss_kl}
\end{equation}
Based on (\ref{eq:gauss_kl}), we define the distance between two risk distributions $r(\mathcal{D}_0)$ and $r(\mathcal{D}_1)$ as
\begin{equation}
\footnotesize
\begin{aligned}
&d\big(r(\mathcal{D}_0),r(\mathcal{D}_1)\big)
=D_{KL}\big(\mathcal{N}_0||\mathcal{N}_1\big)+D_{KL}\big(\mathcal{N}_1||\mathcal{N}_0\big)\\
&=\frac{1}{2}\big(\frac{(\hat{\mu}_{0}-\hat{\mu}_{1})^2+\hat{\sigma_{0}}^2}{\hat{\sigma}_{1}^2}+\frac{(\hat{\mu}_{1}-\hat{\mu}_{0})^2+\hat{\sigma}_{1}^2}{\hat{\sigma}_{0}^2}-2\big).
\vspace{-1mm}
\end{aligned}
\end{equation}
Similar to (\ref{eq:dist1}), the distance $d$ is non-negative and approaches zero when the means and variances from $\mathcal{D}_0$ and $\mathcal{D}_1$ are equal, i.e., $\hat{\mu}_{0}=\hat{\mu}_{1}$ and $\hat{\sigma}_{0}^2=\hat{\sigma}_{1}^2$.

Note that the distance $d$ in both approaches discussed in this section are differentiable, which can facilitate the classifier optimization using gradient-based methods.

\paragraph{Fairness Regularizer $E_{f}$} ~

For simplicity, we assume that the sample set $\mathcal{D}$ has a binary protected attribute, i.e., $|A|=2$. To distinguish the DP and EO cases, we name the regularizers as $E_{f,DP}$ and $E_{f,EO}$, respectively. For TIDP case, $\mathcal{D}$ is splited into two groups: negative protected attribute $\mathcal{D}_{0}$ and positive protected attribute $\mathcal{D}_{1}$. The regularizer fir this case $E_{f,DP}$ is defined as
\vspace{-1mm}
\begin{equation}
    E_{f,DP}(\mathcal{D})=d(r(\mathcal{D}_{0}),r(\mathcal{D}_{1})).
    \vspace{-1mm}
\end{equation}
For TIEO case, $\mathcal{D}$ is splited into four groups: negative and positive samples with negative protected attribute, $\mathcal{D}_{0,n}$ and $\mathcal{D}_{0,p}$, respectively, and negative and positive samples with positive protected attribute, $\mathcal{D}_{1,n}$ and $\mathcal{D}_{1,p}$, respectively. The regularizer for this case $E_{f,EO}$ is defined as
\begin{equation}
\vspace{-1mm}
E_{f,EO}(\mathcal{D})=\sum_{y\in\{n,p\}}d\big(r(\mathcal{D}_{0,y}),r(\mathcal{D}_{1,y})\big).
\vspace{-1mm}
\end{equation}
Note that the regularizer can be extended to the case of multiple protected attributes, i.e., $|A|>2$.

\subsection{CLASSIFIER FORMULATION}
\label{sec:formulation}
As a proof-of-concept, we consider two kinds of classifiers: logistic regression (LR) and support vector machine (SVM). It is worthwhile to note that our proposed method can be easily extended to other complex differentiable classifiers. Define the classifier $g(x)=w^\textup{T}x+b$, and the risk score is computed as $s(x)=\sigma(g(x))$.

{\bf LR Model:} The loss function $L_{\textup{LR}}$ for logistic regression can be expressed as
\vspace{-2mm}
\begin{equation}
\footnotesize
L_{\textup{LR}}(w,b)=\frac{1}{|\mathcal{D}|}\sum_{(X,Y)\in \mathcal{D}}L_{ce}(s(X),Y)+\eta E_f(\mathcal{D}),
\vspace{-2mm}
\label{eqn:lr_loss}
\end{equation}
where $L_{ce}(\cdot)$ denotes the cross-entropy loss of the risk score and the label, and $\eta$ is a positive hyperparameter. Since (\ref{eqn:lr_loss}) is differentialable, we can use gradient-based methods to minimize (\ref{eqn:lr_loss}) with respect to $w$ and $b$.

{\bf Linear SVM Model:} Pegasos was proposed to solve SVM with the hinge loss version using a sub-gradient solver~\citep{shalev2011pegasos}, i.e.,
\vspace{-1mm}
\begin{equation}
\footnotesize
L_{\textup{Pegasos}}(w,b)=\frac{\lambda}{2}||w||^2+\frac{1}{|\mathcal{D}|}\sum_{(X,Y)\in\mathcal{D}}\max\{0,1-Y\cdot g(X)\},
\label{eqn:svm1}
\end{equation}
where $\lambda$ is a positive hyperparameter, and the label $Y\in\{-1,1\}$ in an SVM denotes negative/positive label.

By incorporating the fairness regularizer into (\ref{eqn:svm1}), we obtain the loss function as
\vspace{-1mm}
\begin{equation}
\footnotesize
\begin{aligned}
L_{\textup{LSVM}}(w,b)&=\frac{\lambda}{2}||w||^2+\eta E_f(\mathcal{D})\\
&+\frac{1}{|\mathcal{D}|}\sum_{(X,Y)\in\mathcal{D}}\max\{0,1-Y\cdot g(X)\},
\end{aligned}
\vspace{-1mm}
\label{eqn:svm2}
\end{equation}
where $\lambda$ and $\eta$ are positive hyperparameters. Similar to Pegasos in (\ref{eqn:svm1}), we can apply sub-gradient solvers, due to the sub-differentiability of (\ref{eqn:svm2}).

{\bf Kernel SVM Model:} In a kernel SVM, we use a mapping function $\phi(\cdot)$ to transform the original features to a higher dimensional subspace. Substituting original features $X$ with the mapped features $\phi(X)$, we can rewrite (\ref{eqn:svm2}) as
\begin{equation}
\footnotesize
\begin{aligned}
\hspace{-1.5mm}
L_{\textup{KSVM}}(w,b)&=\frac{1}{|\mathcal{D}|}\sum_{(X,Y)\in\mathcal{D}}\max\{0,1-Y\cdot g(\phi(X))\}\\
&+\frac{\lambda}{2}||w||^2+\eta E_f(\phi(\mathcal{D})).
\end{aligned}
\label{eqn:svm3}
\end{equation}

\begin{figure*}[t]
\centering
\begin{minipage}[t]{1.0\textwidth}
\centering
\includegraphics[width=0.5\linewidth]{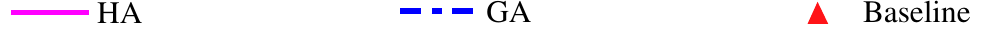}
\end{minipage}
\vskip .1pc
\vfill
\begin{minipage}[t]{0.33\textwidth}
\centering
\includegraphics[width=0.95\linewidth]{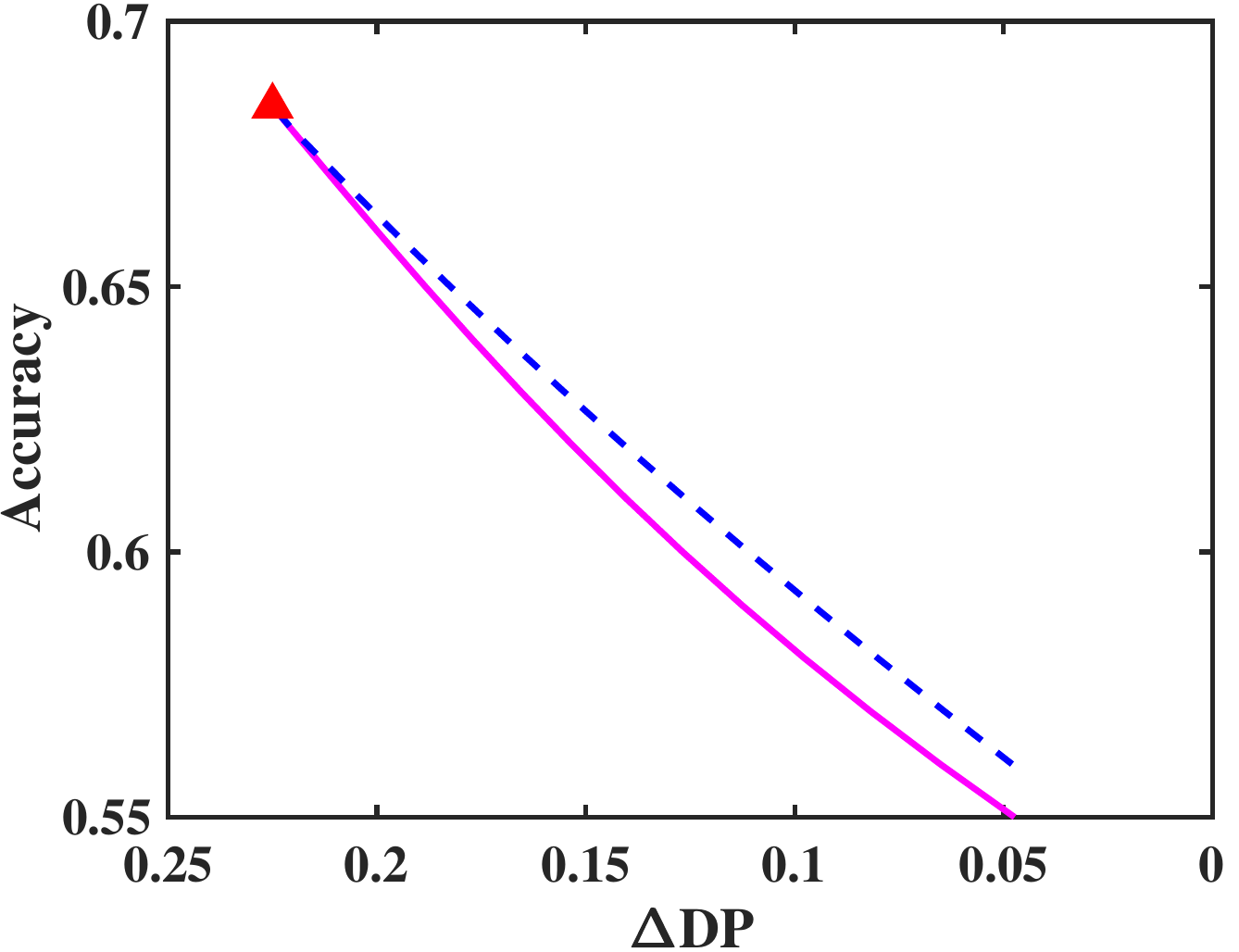}
\centerline{\footnotesize{~~~~~~(a)~LR with DP}}
\end{minipage}
\begin{minipage}[t]{0.33\textwidth}
\centering
\includegraphics[width=0.95\linewidth]{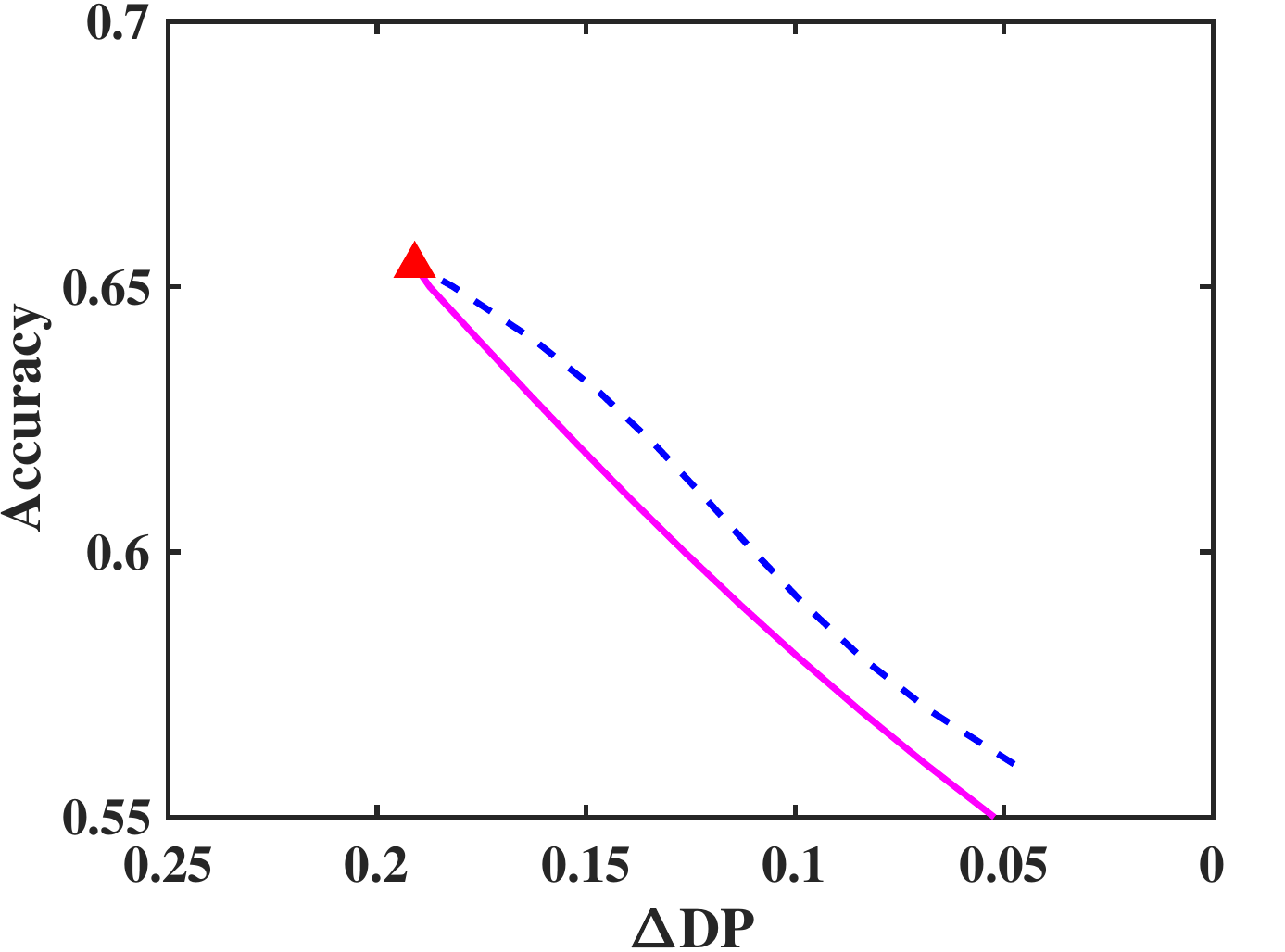}
\centerline{\footnotesize{~~~~~~(b)~LSVM with DP}}
\end{minipage}
\begin{minipage}[t]{0.33\textwidth}
\centering
\includegraphics[width=0.95\linewidth]{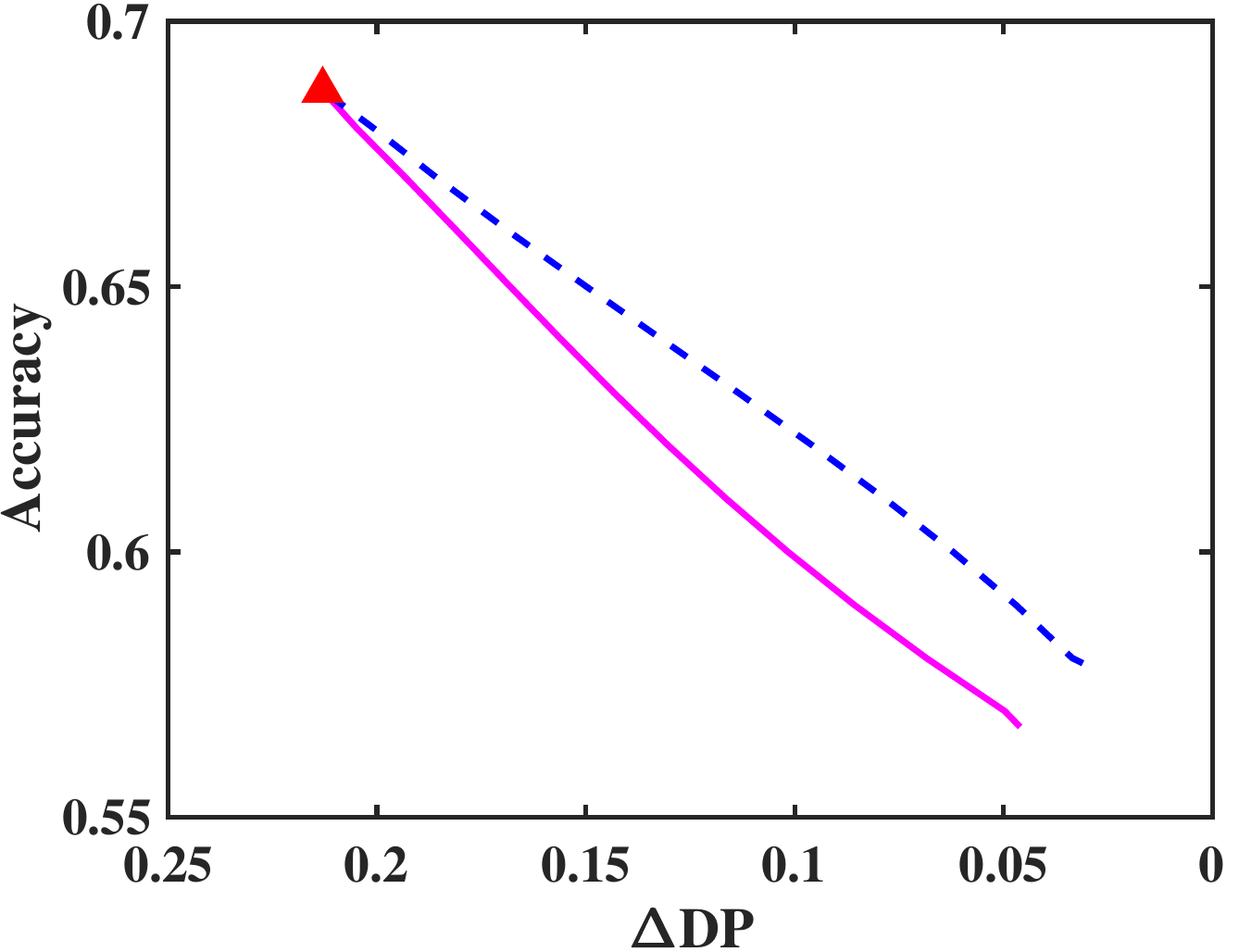}
\centerline{\footnotesize{~~~~~~(c)~KSVM with DP}}
\end{minipage}
\vskip .3pc
----------------------------------------------------------------------------------------------------------------------------------------
\vfill
\begin{minipage}[t]{0.33\textwidth}
\centering
\includegraphics[width=0.95\linewidth]{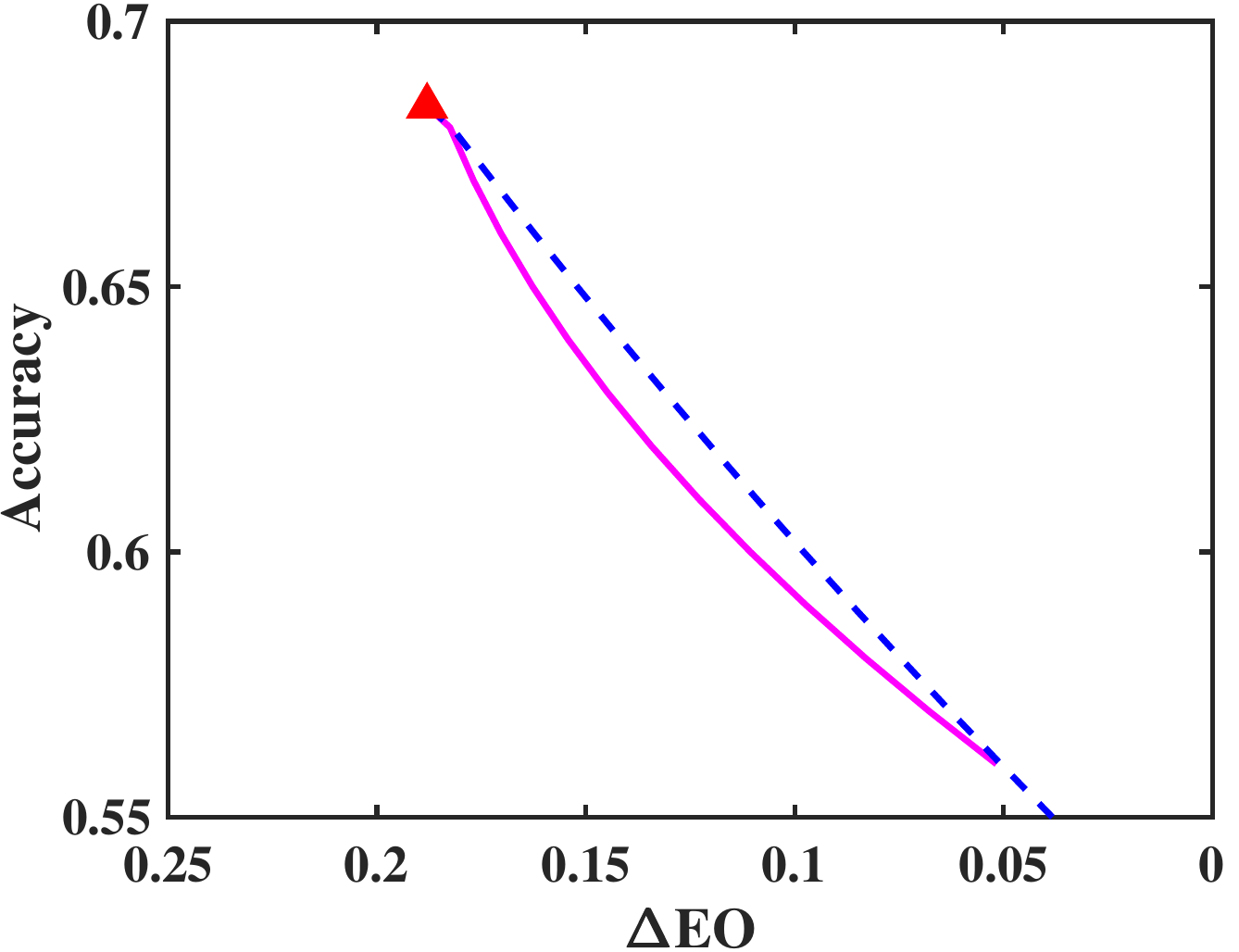}
\centerline{\footnotesize{~~~~~~(d)~LR with EO}}
\end{minipage}
\begin{minipage}[t]{0.33\textwidth}
\centering
\includegraphics[width=0.95\linewidth]{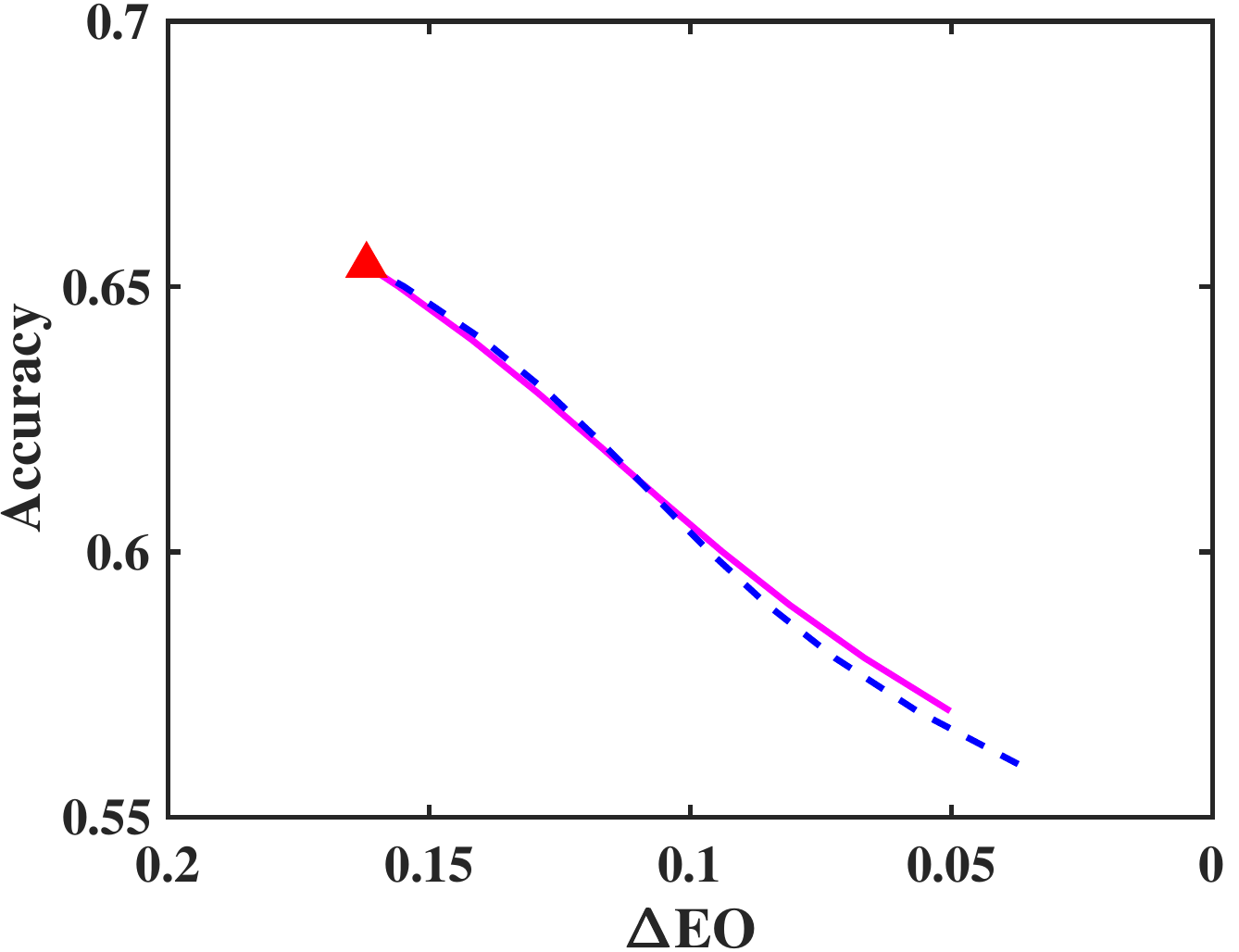}
\centerline{\footnotesize{~~~~~~(e)~LSVM with EO}}
\end{minipage}
\begin{minipage}[t]{0.33\textwidth}
\centering
\includegraphics[width=0.95\linewidth]{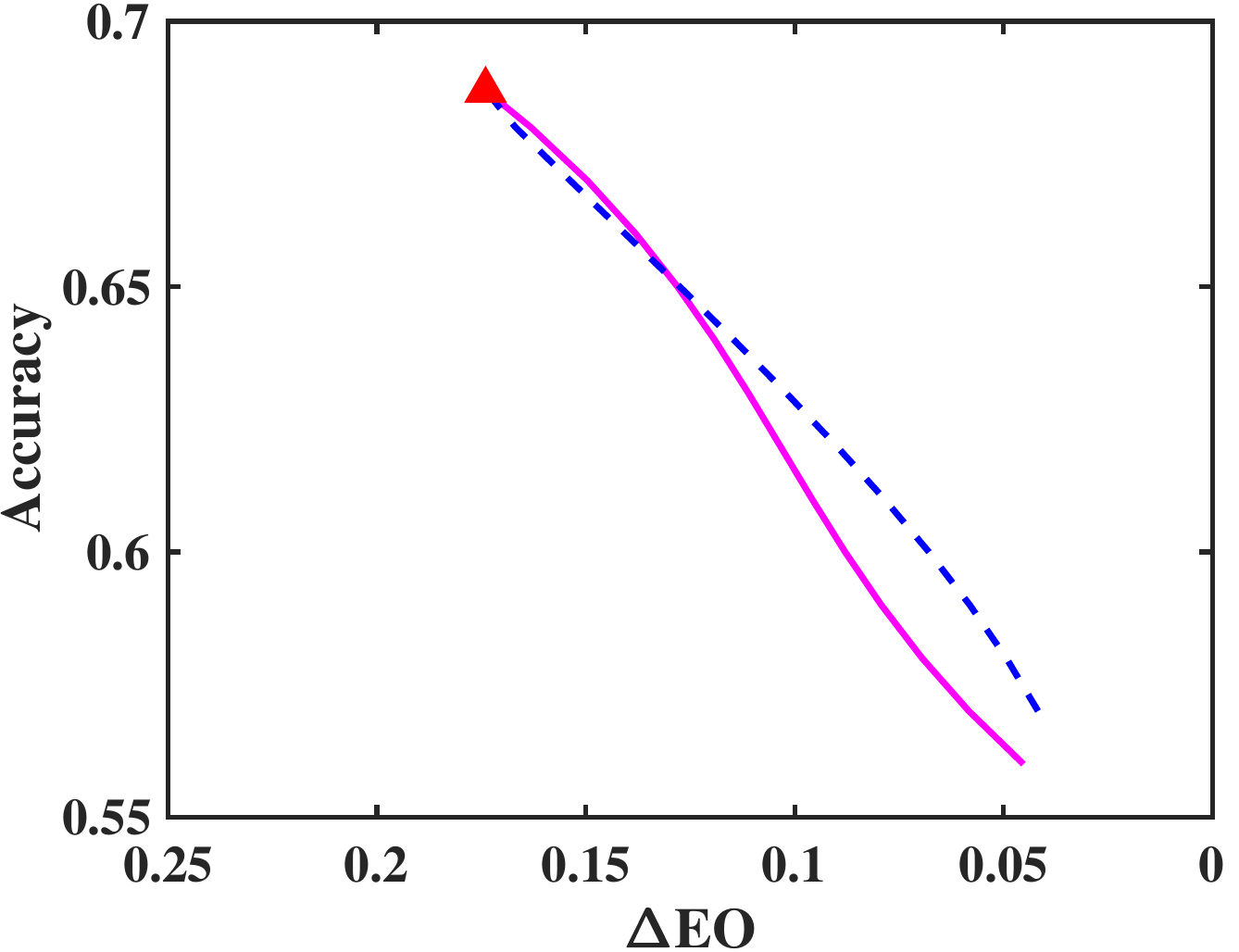}
\centerline{\footnotesize{~~~~~~(f)~KSVM with EO}}
\end{minipage}
\caption{Tradeoff between model accuracy and classification parity index: (a) LR with the DP constraint, (b) LSVM with the DP constraint, (c) KSVM with the DP constraint, (d) LR with the EO constraint, (e) LSVM with the EO constraint, and (f) KSVM with EO constraint. The red filled triangles indicate the baseline cases in each classifier.
}
\label{fig:tradeoff}
\end{figure*}
According to the Representer Theorem~\citep{kimeldorf1971some}, the optimal solution of (\ref{eqn:svm3}) can be spanned by the training samples, i.e., it is of the form $w=\sum_{i=1}^{|\mathcal{D}|}\alpha_i \phi(X_i)$. Hence, the kernel SVM predictor becomes $\tilde{g}(x)=\sum_{i=1}^{|\mathcal{D}|}\alpha_i \phi(X_i)^\textup{T}\phi(x)+b=\sum_{i=1}^{|\mathcal{D}|}\alpha_i K(X_i,x)+b$, where we define the kernel operator $K(X_i,x)=\phi(X_i)^{\textup{T}}\phi(x)$. Equation~(\ref{eqn:svm3}) can be rewritten with respect to $\alpha$ as
\begin{equation}
\footnotesize
\begin{aligned}
\hspace{-1.5mm}
L_{\textup{KSVM}}(\alpha,b)&=\frac{\lambda}{2}\sum_{i,j=1}^{|\mathcal{D}|}\alpha_i\alpha_j K(X_i,X_j)+\eta E_f(\phi(\mathcal{D}))\\
&+\frac{1}{|\mathcal{D}|}\sum_{(X,Y)\in \mathcal{D}}\max\{0,1-Y\cdot \tilde{g}(X)\},
\label{eqn:svm4}
\end{aligned}
\end{equation}
where $\lambda$ and $\eta$ are positive hyperparameters. To calculate the fairness regularizer, we can use the kernel trick to compute the risk scores and the risk distributions over the mapped sample set $\phi(\mathcal{D})$. Similarly, we can use subgradient-based methods to optimize the loss function (\ref{eqn:svm4}) with respect to $\alpha$.

\section{EXPERIMENTS}
In this section, we carry out experiments on the COMPAS risk assessment dataset compiled by ProPublica~\citep{larson2016we} and evaluate how well our proposed method can alleviate the problem of threshold variant fairness in the prior art of classification parity.

\subsection{DATASET AND FEATURE PROCESSING}
ProPublica compiled a list of all criminal defendants screened through the COMPAS tool in Broward County, Florida, during 2013 to 2014.\footnote{The dateset can be downloaded from \url{https://github.com/propublica/compas-analysis}.} The dataset contains defendants' records of prison times, demographics (e.g., gender, race, and age), criminal histories (current charge type, charge degree, and number of prior crimes), the COMPAS risk scores, and the ground truth of recidivism within two years after the screening. Previous discussions on bias decisions from machine learning on this dataset can be found in \citep{angwin2016machine,dieterich2016compas,dressel2018accuracy}.

For simplicity, we only consider a subset of defendants whose race was either African-American or Caucasian, which is the protected attribute in our study. The set of features used in the classification task is summarized in Table~\ref{tab:compas}. We only use the protected attribute, ``race'', as an indicator of the groups in the training and exclude it from the features for the classification task. The features are a combination of continuous and categorical features. For continuous features, we subtract their mean and scale them to unit variance; for categorical features, we use $0/1$ to encode the features, expect for the labels in SVM where $-1/+1$ encoding is used. The dataset was randomly split into training ($70\%$) and test ($30\%$).
\begin{table}[t]
\caption{Preprocessed Features Used in COMPAS Data}
\begin{center}
\begin{tabular}{l|ll}\hline
\multicolumn{1}{c|}{\bf Feature} & \multicolumn{1}{c}{\bf Type} & \multicolumn{1}{c}{\bf Preprocessing} \\ \hline
Age & Continuous & Normalization\\
Gender & Binary  & 0-1 encoding \\
Race (Protected)  & Binary & 0-1 encoding  \\
Priors count & Continuous & Normalization \\
Charge degree & Binary& 0-1 encoding \\ \hline
Recidivism (Label)& Binary& 0-1 encoding \\ \hline
\end{tabular}
\end{center}
\label{tab:compas}
\end{table}
\begin{figure*}[t]
\begin{minipage}[h]{0.03\textwidth}
\centerline{\footnotesize{}}
\end{minipage}
\begin{minipage}[h]{0.46\textwidth}
\centering
\includegraphics[width=0.6\linewidth]{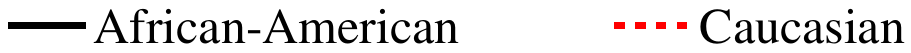}
\end{minipage}
\hspace{2mm}
\begin{minipage}[h]{0.46\textwidth}
\centering
\includegraphics[width=0.75\linewidth]{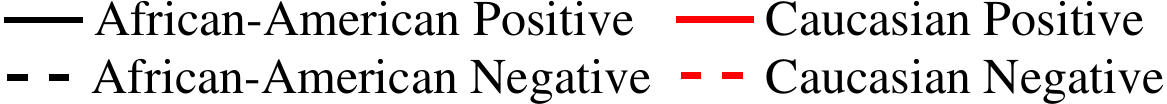}
\end{minipage}
\vfill
\begin{minipage}[h]{0.03\textwidth}
\centerline{\scriptsize{\rotatebox{90}{Unconstrained (baseline)}}}
\end{minipage}
\begin{minipage}[h]{0.46\textwidth}
\centering
\includegraphics[width=0.5\linewidth]{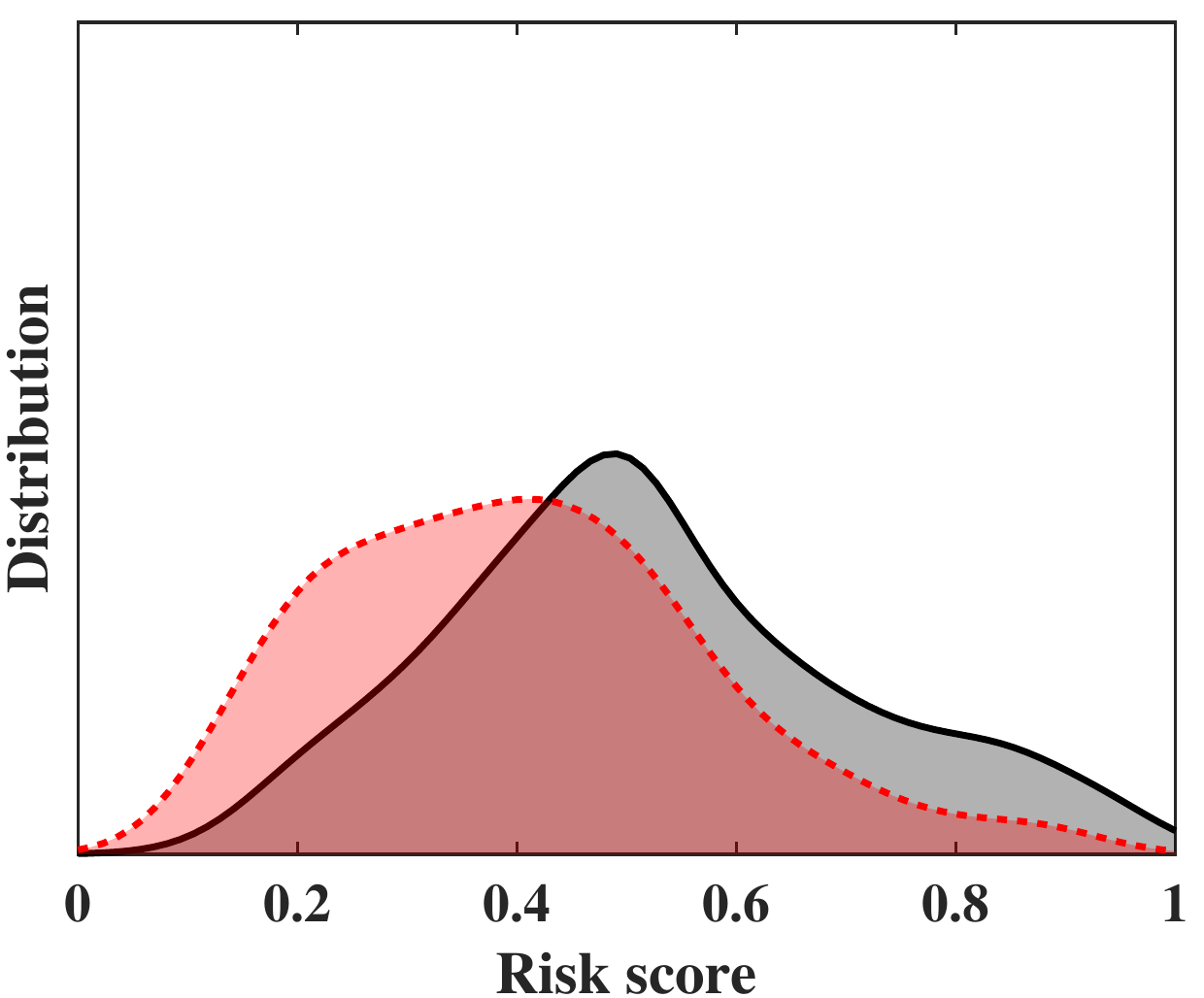}
\end{minipage}
\hspace{0.02mm}
\begin{minipage}[h]{0.01\textwidth}
\centering
\includegraphics[width=0.11\linewidth]{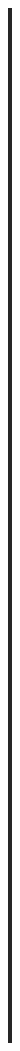}
\end{minipage}
\begin{minipage}[h]{0.46\textwidth}
\centering
\includegraphics[width=0.5\linewidth]{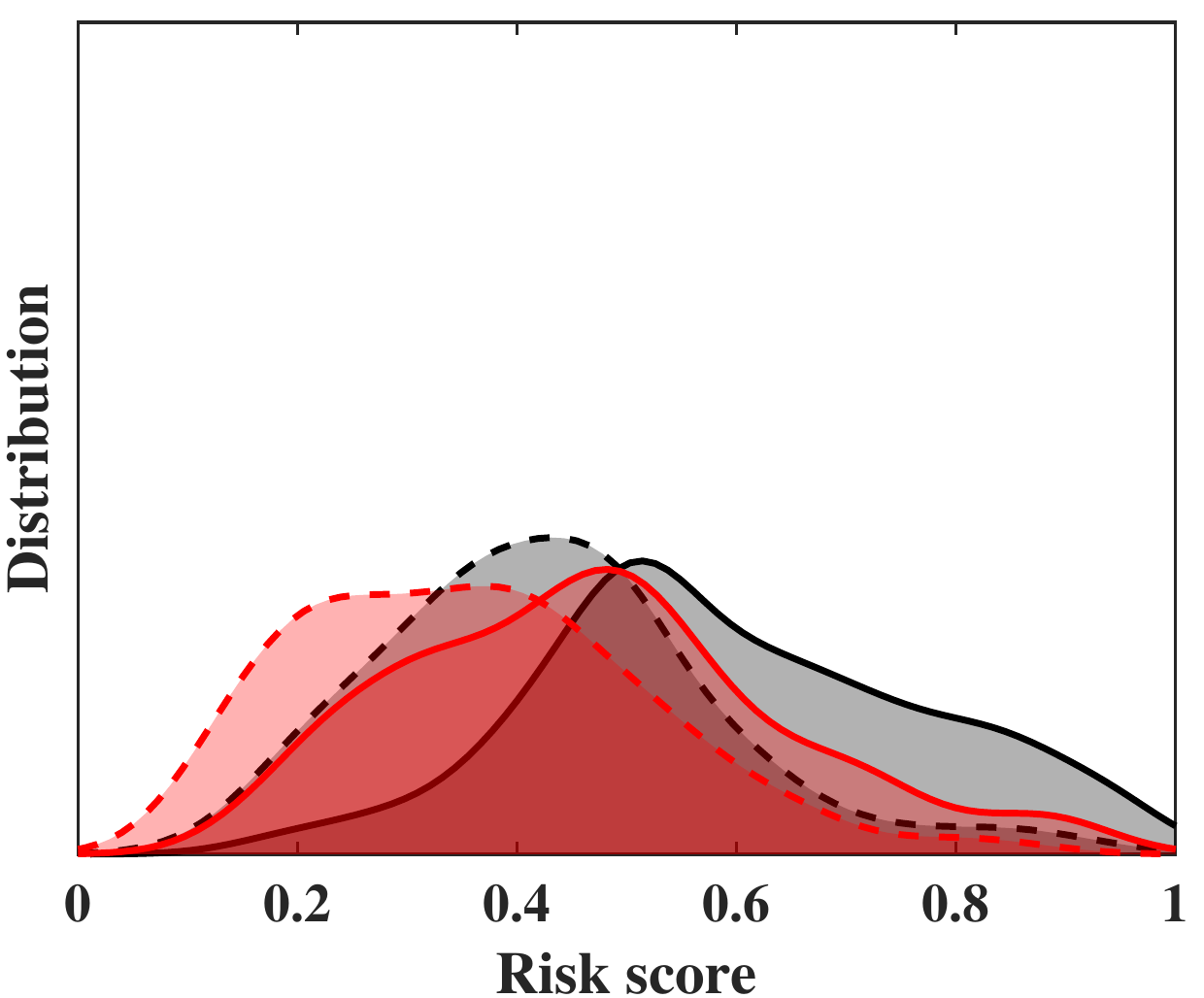}
\end{minipage}
\vfill
\begin{minipage}[h]{0.03\textwidth}
\centerline{\scriptsize{\rotatebox{90}{Moderate constraints}}}
\end{minipage}
\begin{minipage}[h]{0.23\textwidth}
\centering
\includegraphics[width=1.0\linewidth]{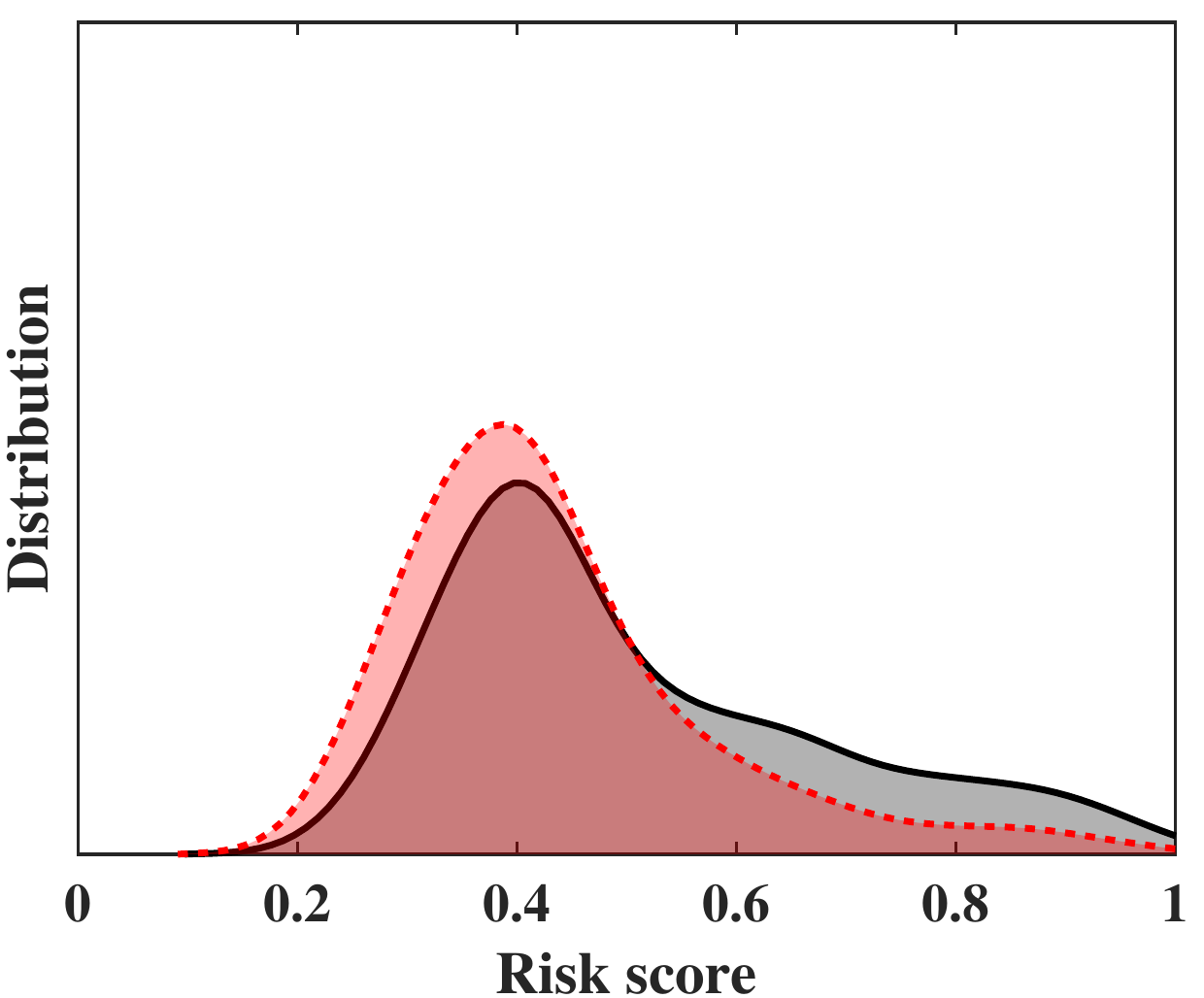}
\end{minipage}
\begin{minipage}[h]{0.23\textwidth}
\centering
\includegraphics[width=1.0\linewidth]{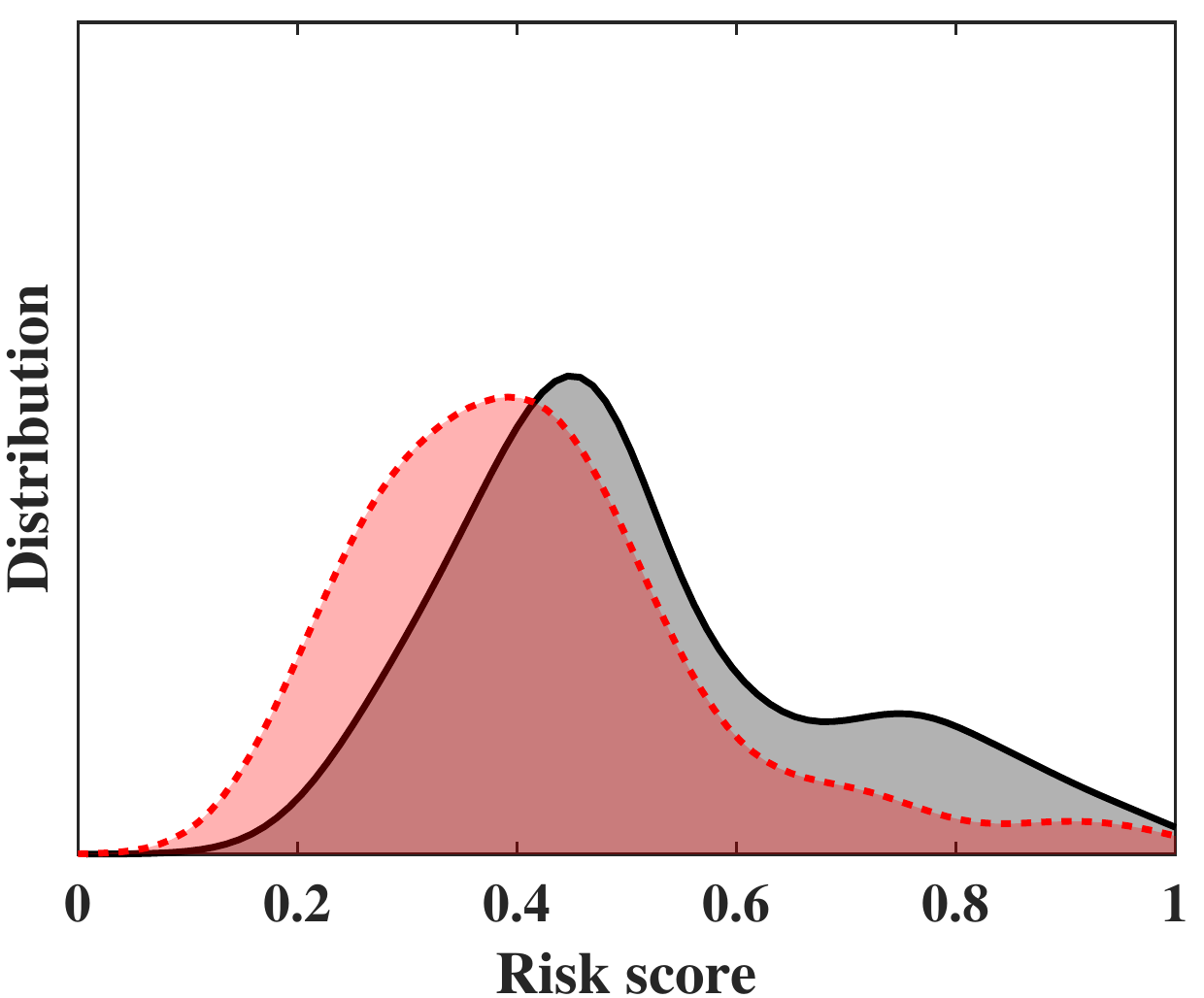}
\end{minipage}
\begin{minipage}[h]{0.01\textwidth}
\centering
\includegraphics[width=0.11\linewidth]{fig/bar.pdf}
\end{minipage}
\begin{minipage}[h]{0.23\textwidth}
\centering
\includegraphics[width=1.0\linewidth]{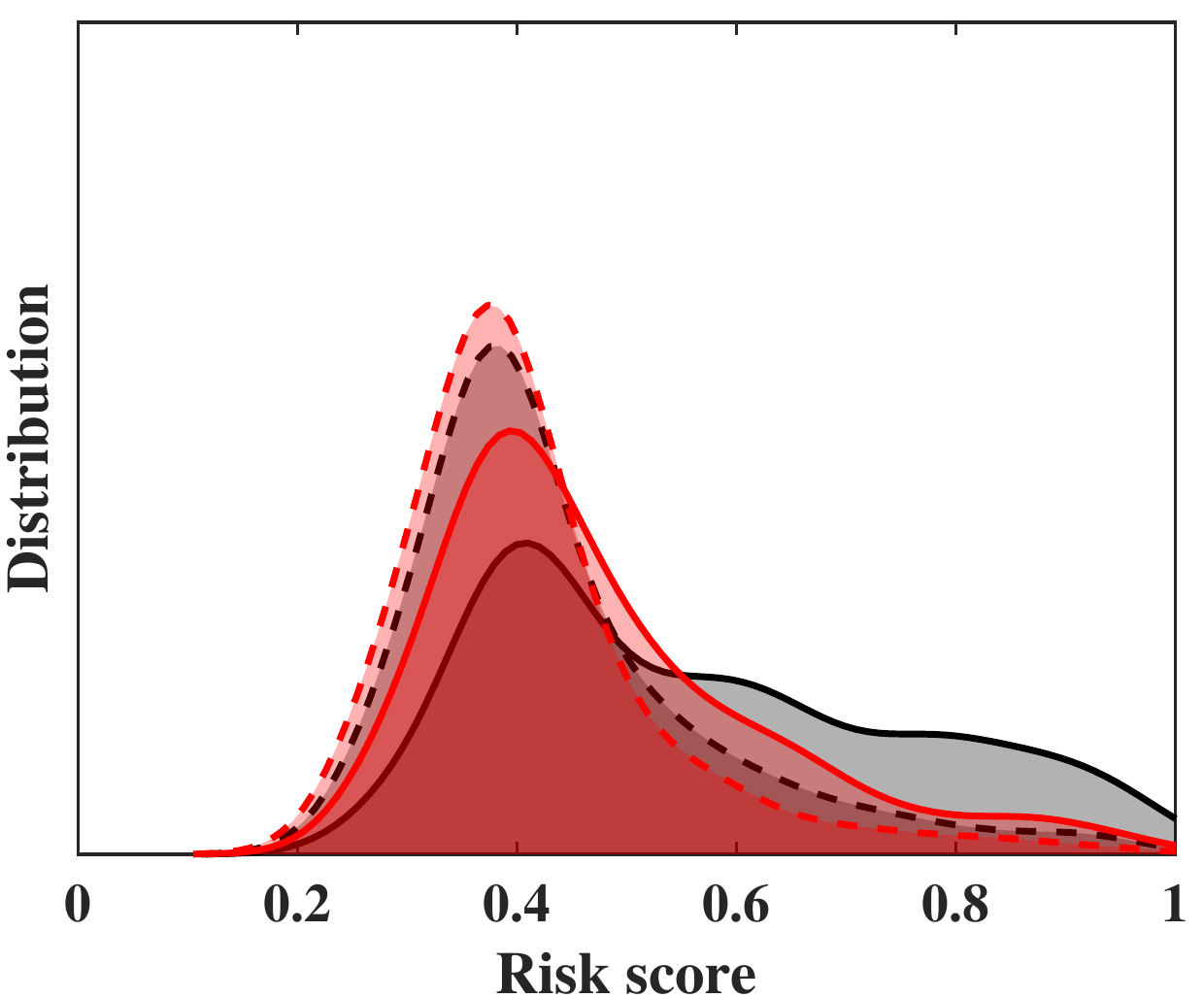}
\end{minipage}
\begin{minipage}[h]{0.23\textwidth}
\centering
\includegraphics[width=1.0\linewidth]{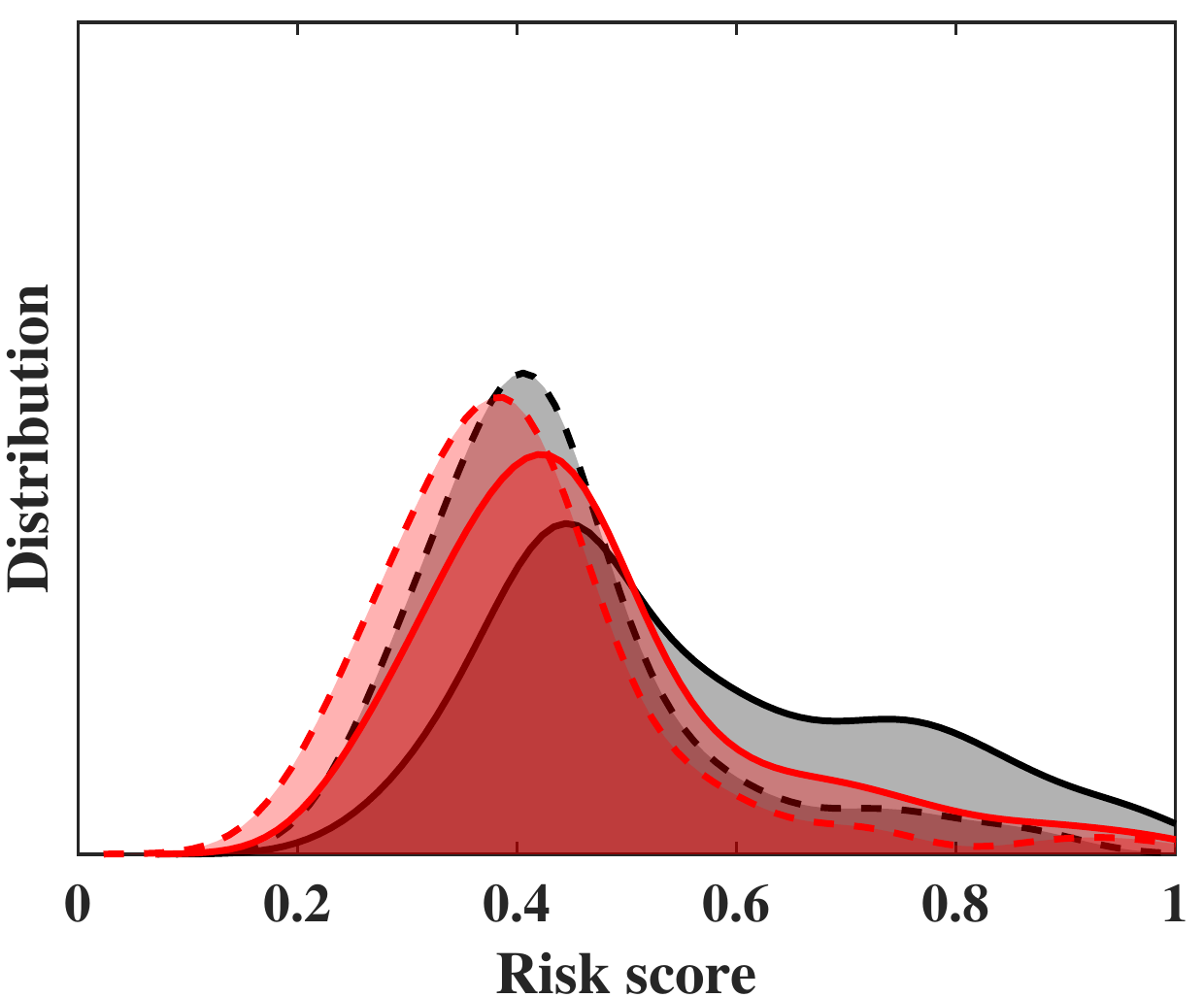}
\end{minipage}
\vfill
\begin{minipage}[h]{0.03\textwidth}
\centerline{\scriptsize{\rotatebox{90}{~~~~~~~~~~~Strong constraints}}}
\end{minipage}
\begin{minipage}[h]{0.23\textwidth}
\centering
\includegraphics[width=1.0\linewidth]{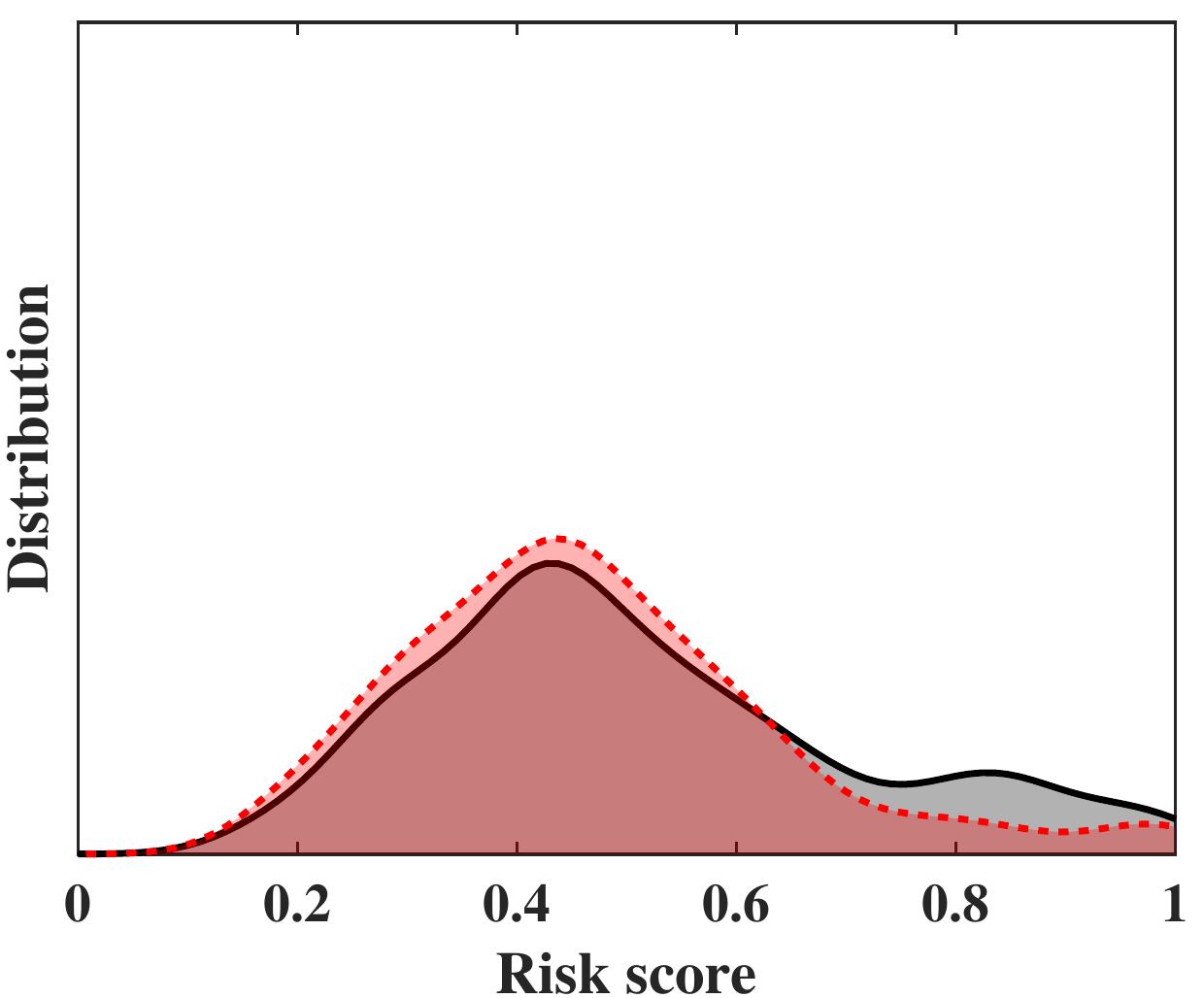}
\end{minipage}
\begin{minipage}[h]{0.23\textwidth}
\centering
\includegraphics[width=1.0\linewidth]{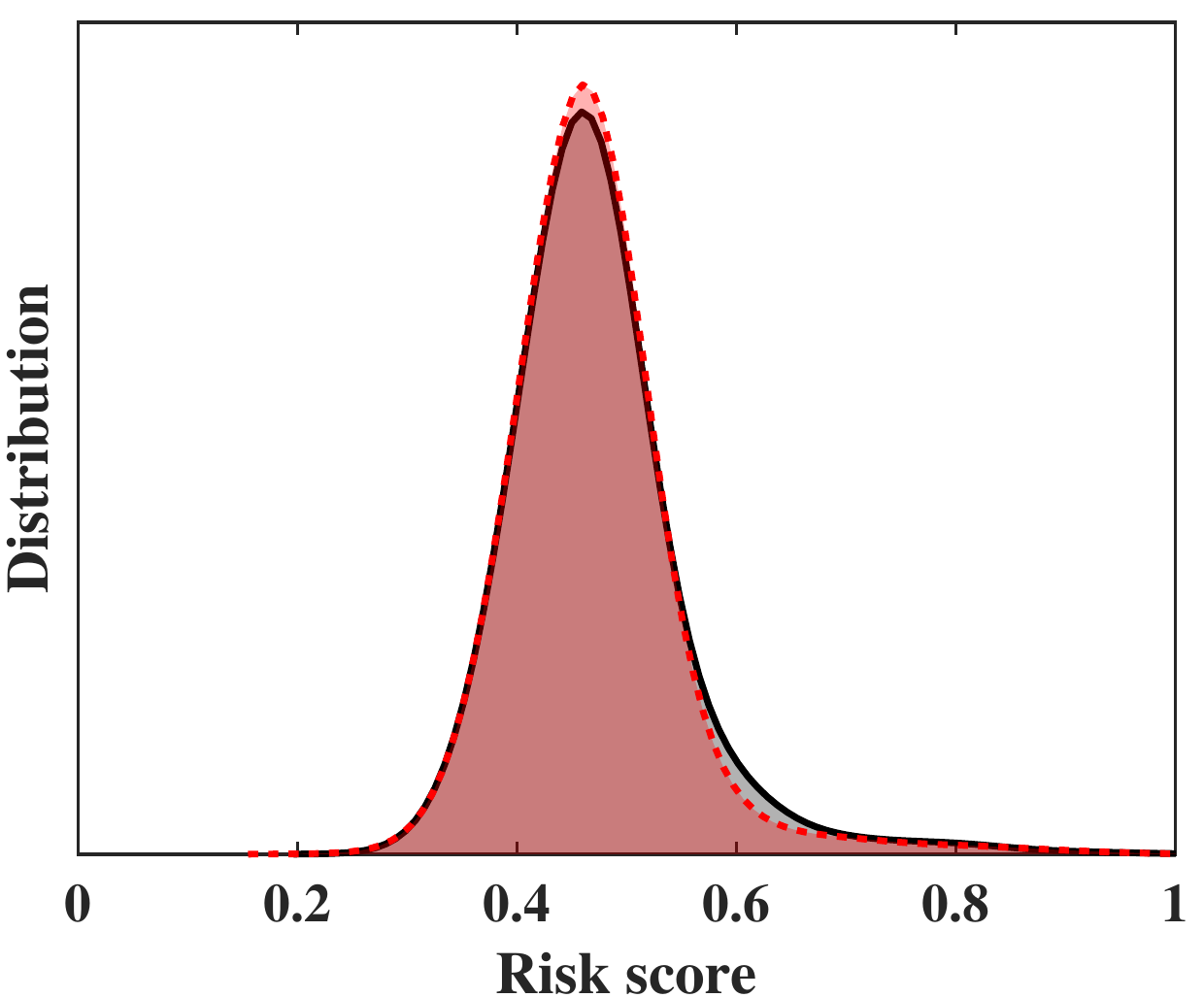}
\end{minipage}
\begin{minipage}[h]{0.01\textwidth}
\centering
\includegraphics[width=0.11\linewidth]{fig/bar.pdf}
\end{minipage}
\begin{minipage}[h]{0.23\textwidth}
\centering
\includegraphics[width=1.0\linewidth]{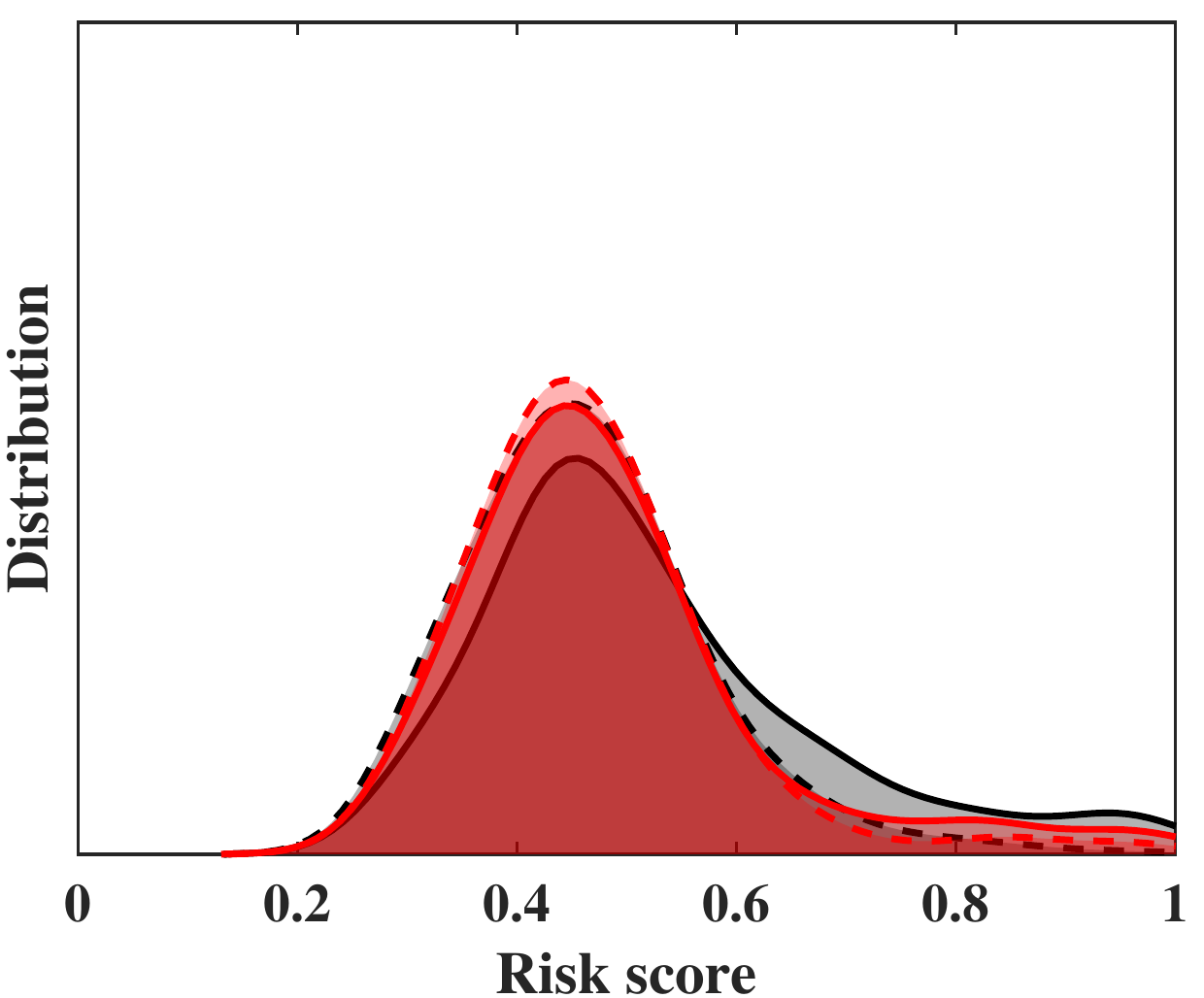}
\end{minipage}
\begin{minipage}[h]{0.23\textwidth}
\centering
\includegraphics[width=1.0\linewidth]{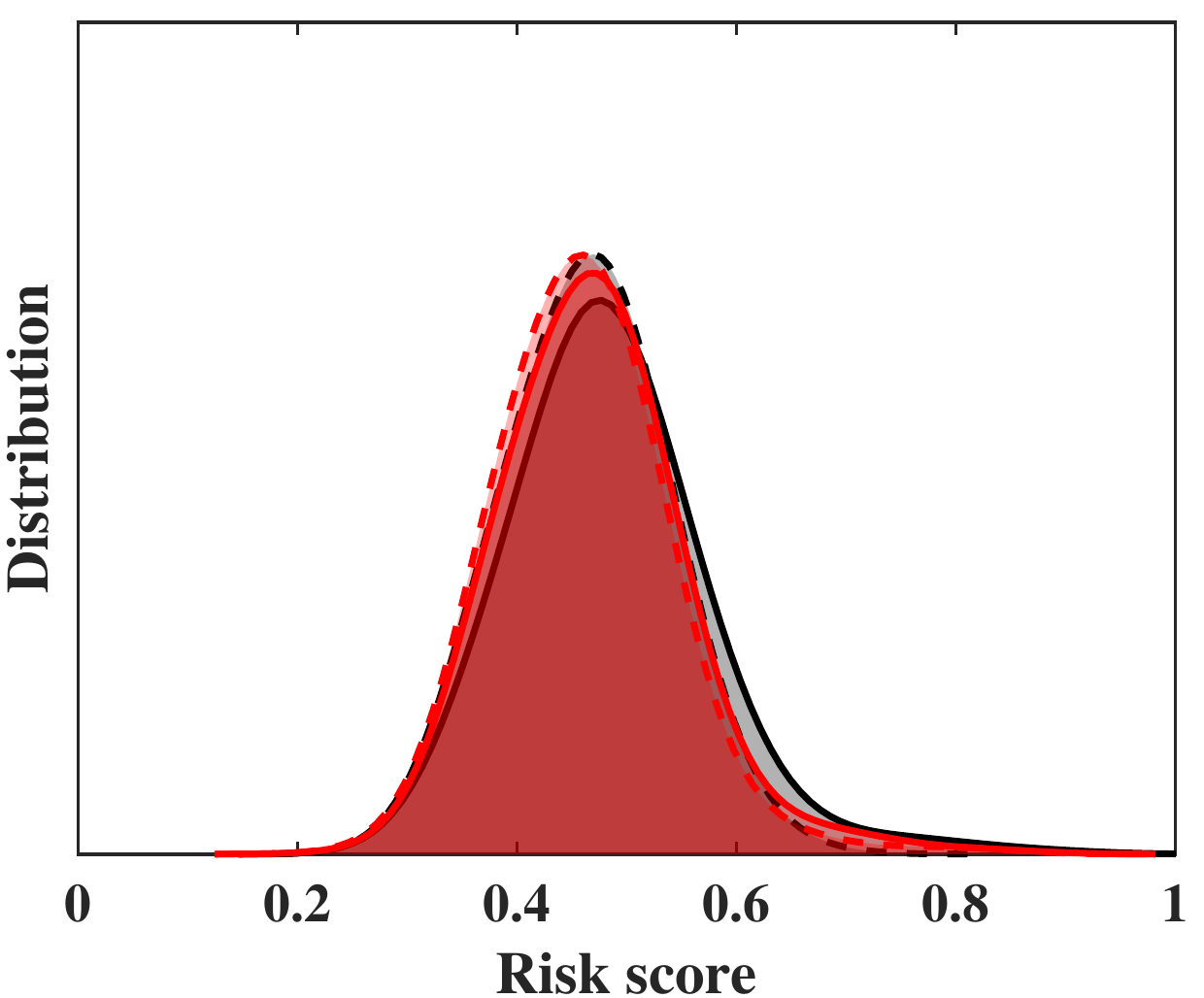}
\end{minipage}
\vfill
\begin{minipage}[h]{0.26\textwidth}
\centerline{\footnotesize{~~~~~~~~(a) DP+HA}}
\end{minipage}
\begin{minipage}[h]{0.23\textwidth}
\centerline{\footnotesize{~~(b) DP+GA}}
\end{minipage}
\begin{minipage}[h]{0.23\textwidth}
\centerline{\footnotesize{~~~~~~~~(c) EO+HA}}
\end{minipage}
\begin{minipage}[h]{0.23\textwidth}
\centerline{\footnotesize{~~~~~~~~(d) EO+GA}}
\end{minipage}
\caption{Risk distributions of LR classifiers with different levels of fairness regularizers. The first row is the baseline, i.e., unconstrained classifier; the second row is the classifier with moderate constraints; the third row is the classifier with strong constraints. (a) DP+HA, (b) DP+GA, (c) EO+HA, and (d) EO+GA. The figures are best viewed in color.}
\label{fig:vis}
\end{figure*}

\subsection{TRADEOFF BETWEEN ACCURACY AND FAIRNESS}
We have trained three types of classifiers: LR, linear SVM (LSVM), and kernel SVM (KSVM). For SVM, we set hyperparameter $\lambda=\frac{1}{10\cdot |\mathcal{D}|}$. We used radial basis function (RBF) kernel with $\gamma=0.5$ in KSVM, i.e., $K(x_1,x_2)=\exp(-\gamma||x_1-x_2||_2^2)$. We applied gradient descent to optimize the models with momentum $0.9$ and learning rate starting from $0.1$ to $0.0001$. 

We tuned $\eta$ from $0$ to $5$ in each classifier to show the tradeoff curve of accuracy versus fairness in Figure~\ref{fig:tradeoff} via the two proposed equalization approaches: histogram approximation (HA) and gaussian assumption (GA). Note that the classifiers with $\eta=0$ are the unconstrained classifiers, considered as {\it baseline}.

As shown in Figure~\ref{fig:tradeoff}, we can see that our proposed methods are effective at alleviating disparate learning performances across the groups. The CV scores of $\Delta$DP and $\Delta$EO reduce from around $0.2$ to $0.05$, with the help of the proposed fairness regularizers, whereas the resulting accuracies drop by approximately $0.1$ on a $[0,1]$ scale. Generally, our proposed methods can be used in different types of classifiers (e.g., LR, LSVM, and KSVM) and different types of classification parity indices (e.g., DP and EO). Comparing two equalization methods of HA and GA, the models with GA based fairness regularizer have slightly better performances than those with the HA approach in terms of the accuracy cost to achieve the same fairness level.

\begin{table*}[t]
\centering
\begin{minipage}[t]{0.45\textwidth}
\caption{Performance Under DP Constraint}
\vspace{-3mm}
\begin{center}
\begin{tabular}{c|cc|cc}\hline
& {\bf Accuracy}  & {\bf $\Delta$DP} & {\bf Int.} &{\bf STD}\\ \hline
Baseline &  68.4\%  & 0.225 & 0.145 & 0.044\\
Zafar-DP &  57.4\%  & 0.060  & 0.079 & 0.024\\\hline
LR-HA  & 56.6\%  & 0.075  & 0.089 & 0.024\\
LR-GA  & 56.9\%  & 0.066  & 0.093 & 0.023\\
LSVM-HA &  56.2\%  & 0.035 & 0.073 & 0.022\\
LSVM-GA &  56.5\%  & 0.048 & 0.097 & 0.020\\
KSVM-HA &  57.0\%  & 0.059 & 0.057 & 0.014\\
KSVM-GA &  58.4\%  & 0.064 & 0.098 & 0.024\\\hline
\end{tabular}
\end{center}
\label{tab:dp}
\end{minipage}
\hspace{8mm}
\begin{minipage}[t]{0.45\textwidth}
\caption{Performance Under EO Constraint}
\vspace{-3mm}
\begin{center}
\begin{tabular}{c|cc|cc}\hline
& {\bf Accuracy}  & {\bf $\Delta$EO} & {\bf Int.} &{\bf STD}\\ \hline
Baseline &  68.4\% & 0.188 &0.123 & 0.037 \\
Zafar-EO &  63.2\% & 0.120 &0.154 & 0.047 \\\hline
LR-HA  & 62.7\% & 0.117 &0.077 & 0.024 \\
LR-GA  & 62.9\% & 0.117 &0.098 & 0.028 \\
LSVM-HA &  63.3\% & 0.137 & 0.109 & 0.029 \\
LSVM-GA &  63.0\% & 0.129 & 0.062 & 0.018 \\
KSVM-HA &  63.2\% & 0.112  & 0.067 & 0.012 \\
KSVM-GA &  62.9\% & 0.096  & 0.040 & 0.009 \\\hline
\end{tabular}
\end{center}
\label{tab:eo}
\end{minipage}
\end{table*}

\subsection{VISUALIZATION OF RISK DISTRIBUTION}

In this section, we use linear regression classifiers as an example to visualize the influence of the proposed fairness regularizers on risk distributions. Figure~\ref{fig:vis} shows the risk distributions of this type of classifiers in three example cases: with no constraints (baseline), medium constraints, and large constraints. For the DP constraint, we present the risk distributions of the African-American group and the Caucasian group; for the EO constraint, we show the risk distributions of African-American recidivists, African-American non-recidivists, Caucasian recidivists, and Caucasian non-recidivists, respectively.

Several observations can be made from Figure~\ref{fig:vis}. First, the distributions from the baseline classifiers are distinct among the groups, suggesting the discriminating decision performances for different groups. The unconstrained classifiers suffer from some bias in their predictions. Second, our proposed methods progressively equalize the risk distributions among the groups as we weigh more on fairness regularizers. When we impose medium constraints, the corresponding risk distributions become closer in shape than that in the baseline cases. When we keep increasing the constraints to a significant level, the corresponding risk distributions almost overlap with each other, indicating the effectiveness of the regularizer to equalize the risk distributions. 

\subsection{PARITY VARIATION VERSUS THRESHOLD}

We conducted experiments on three types of classifiers (e.g., LR, LSVM, and KSVM) with 2 equalization approaches (e.g., HA, GA). In this section, we name the proposed classifiers as ``\{classifier type\}-\{equalization\}''. For instance, ``LR-HA'' refers to LR classifier with HA approach. We compare our proposed approaches with the prior works based on DP constraint~\citep{zafar2017dp} and EO constraint~\citep{zafar2017eo}. We refer to them as ``Zafar-DP'' and ``Zafar-EO'' in the paper, respectively. The baseline refers to the unconstrained LR classifier. Note that all these methods do not use protected attributes to make predictions in the test stage. We show the variation of classification parity versus the decision threshold of the risk score for a given classifier. For fair comparison, we tune the parameter $\eta$ to make the proposed classifiers have comparable performances to the prior works, in terms of prediction accuracy. Performance accuracy and the CV scores of $\Delta$DP and $\Delta$EO come from the classifiers with the default decision threshold.

\begin{figure*}[t]
\centering
\begin{minipage}[h]{0.475\textwidth}
\includegraphics[width=1.0\linewidth]{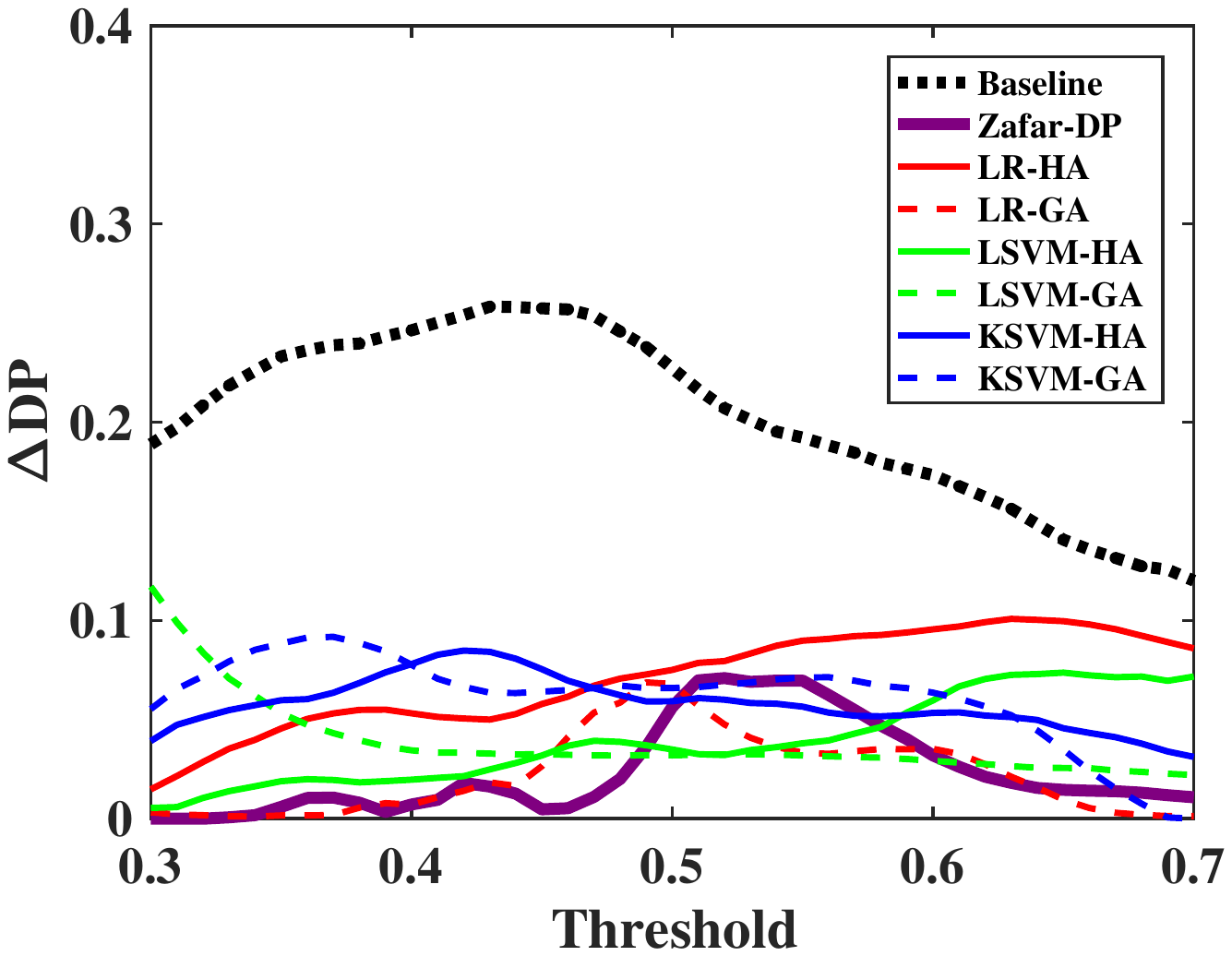}
\centerline{\footnotesize{~~~~~~~~~(a)}}
\end{minipage}
\hspace{3mm}
\begin{minipage}[h]{0.475\textwidth}
\includegraphics[width=1.0\linewidth]{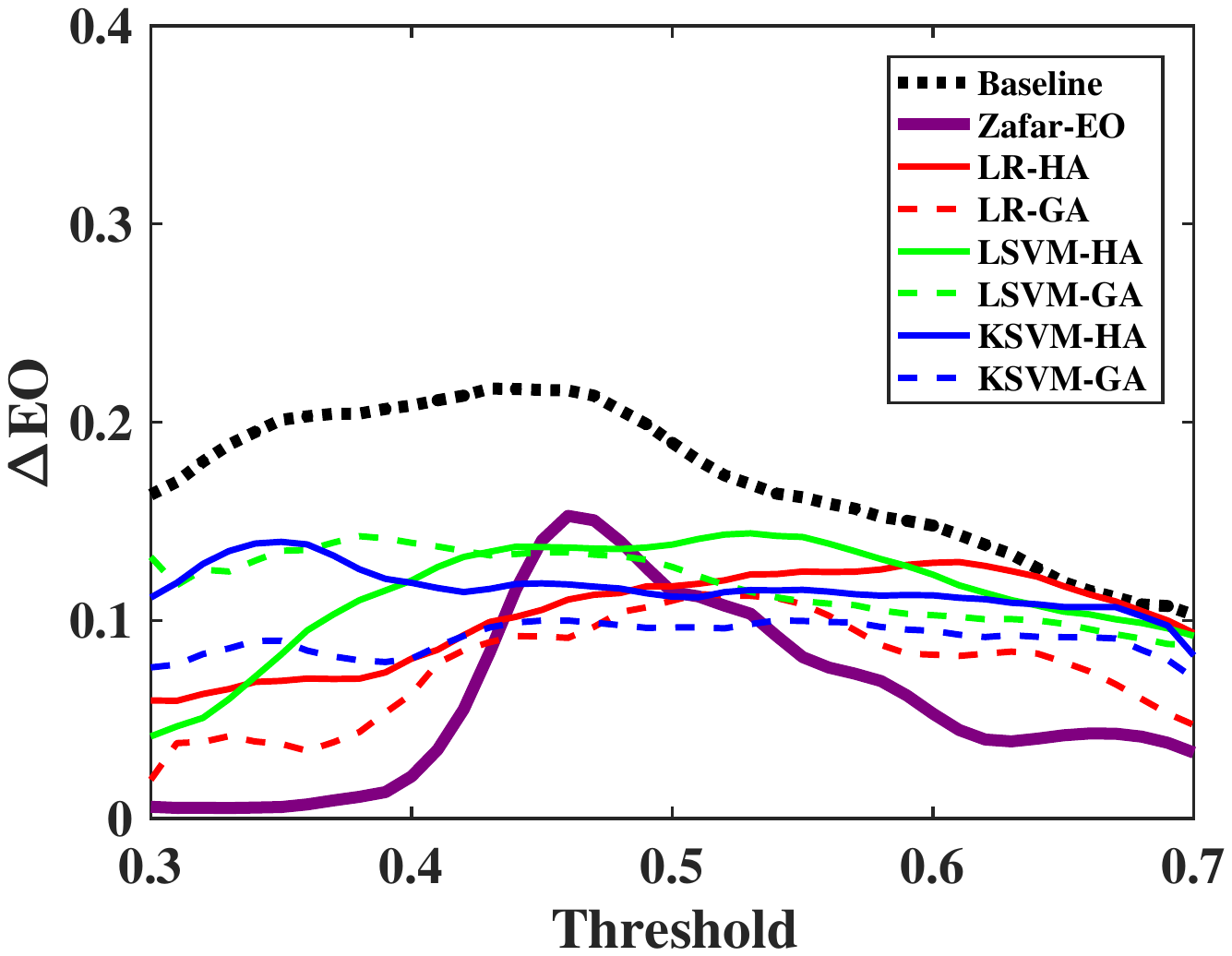}
\centerline{\footnotesize{~~~~~~~~~(b)}}
\end{minipage}
\caption{Variations of classification parity versus decision threshold: (a) classifiers with the DP constraint and (b) classifiers with the EO constraint. The figures are best viewed in color.}
\label{fig:curve}
\end{figure*}

To investigate the parity variation versus the decision threshold, we present the variation of the classification parity by changing the decision threshold ranging from $0.3$ to $0.7$. Table~\ref{tab:dp} presents the performances of the classifiers with DP constraint and Table~\ref{tab:eo} presents the performances of the classifiers with EO constraint. ``Int.'' refers to the interval length of the parity variation and ``STD'' denotes the standard deviation. Figure~\ref{fig:curve} illustrates the variations of classification parity versus the risk score threshold ranging from $0.3$ to $0.7$. Less fluctuation and more flatness of the curve suggest that the classifier achieves higher threshold-invariant parity.

As we expect, the baseline classifier has the best performance in terms of accuracy ($68.4\%$), but worst in CV scores of classification parity ($\Delta$DP$=0.225$ and $\Delta$EO$=0.188$). It spans a wider range and has a larger variance in parity when decision threshold ranges from $0.3$ to $0.7$. The variation curves in Figure~\ref{fig:curve} are also above the classifiers with fairness constraints. With the fairness constraints during the training, the CV scores are both reduced in Table~\ref{tab:dp} and \ref{tab:eo} at the cost of the accuracy performances, and variation ranges and variances decrease to a lower level, indicating the better equitable decision performances among the groups in terms of classification parity. For the DP constraint in Table~\ref{tab:dp}, the proposed classifiers have slightly better DP consistency against the change of decision threshold than Zafar-DP, according to the SD of $\Delta$DP. For the EO constraint in Table~\ref{tab:eo}, the proposed classifiers have a noticeable advantage in invariant EO against the change of decision threshold. Compared with Zafar-EO, the proposed classifiers have smaller interval ranges and smaller variances of $\Delta$EO.

From Figure~\ref{fig:curve}, we can observe that our proposed methods have minor fluctuations on the classification parity curves, compared with Zafar-DP~\citep{zafar2017dp} and Zafar-EO~\citep{zafar2017eo}, respectively. Slightly modifying the decision threshold do not affect the classification parity to a large extent in the proposed classifiers. This suggests that our proposed constraints are effective in reducing the sensitivity of classifiers to the decision threshold in terms of classification parity. The noticeable drops in the CV scores in the three types of the classifiers demonstrate that our methods can be applied to a variety of classifiers.

\subsection{DISCUSSIONS}
From the experimental results, we can see that two proposed methods are effective in equalizing the risk distributions so as to achieve threshold invariant fairness. In this section, we discuss and summarize the difference between the two methods.

The HA method uses an $N$-bin histogram to compute the risk distribution. To make the histogram operation differentiable, a Gaussian kernel is used to approximate the accumulation process in each histogram bin. The HA method requires two hyperparameters: the number of bins $N$ and the variance $\sigma_c^2$ in (\ref{eq:gauss}). Their settings influence the computational cost and the precision of the distribution approximation. The HA method computes the weight of every sample to each histogram bin via a Gaussian kernel and then sum up the weights in each bin to construct the risk distribution. The overall computational complexity of computing risk distribution in HA is $O(MN)$, where $M$ is the number of training samples.

The GA method assumes the Gaussian distribution and uses only the statistics of the first and second moments (mean and variance) to characterize the risk distribution. 
The estimation of means and variances has a computational complexity of $O(M)$, which is lower than the HA method. Considering the similar performances between the two methods and the difference in computational complexities, we can see that the GA method is a preferred approach to achieve the equalization of risk distributions.

In addition, our proposed methods can be extended to the case of multiple protected attributes and other complex differentiable classifiers. For example, the proposed methods are compatible with neural network classifiers. To make our methods adaptive to the batch-based training of neural networks, we can modify gradient descent optimizer to batch-based optimizer (e.g., stochastic gradient descent) and calculate the risk distributions over one training batch instead of over the whole training set.

\section{CONCLUSION}
In this paper, we have introduced a novel fairness notion in machine learning models. Different from the prior definition of classification parity, the new notion aims to build fair classifiers that are not sensitive to the decision thresholds, and achieves this by equalizing the risk distributions among different groups. We perform distribution equalization using histogram approximation and Gaussian assumption, respectively, and incorporate them into logistic regression and SVM classifiers. Experimental results show that our fairness notion allows better control of the fairness level against the decision threshold of the classifier, and has broad compatibility to multiple types of machine learning classifiers.

\renewcommand{\baselinestretch}{1}
\bibliographystyle{apa-good}
\bibliography{sample}

\newpage
\onecolumn
\begin{appendix}
\section{Relation of Equalized Odds and Risk Distributions}
We denote $P_{a,y}(\hat{Y}):=P(\hat{Y}|A=a,Y=y)$ and $f_{a,y}(s):=f(s|A=a,Y=y)$ as the risk score of one sample and risk distributions over the group with protected attribute $A$ and label $Y$, respectively. Given the decision threshold $t\in[0,1]$, we have
\begin{equation}
\begin{aligned}
\Delta \textup{EO}&=\frac{1}{2}\Big(\big |P_{0,0}(\hat{Y}=1)-P_{1,0}(\hat{Y}=1)\big |+\big |P_{0,1}(\hat{Y}=1)-P_{1,1}(\hat{Y}=1)\big|\Big)\\
&=\frac{1}{2}\Big(\big|\int_{t}^{1} f_{0,0}(s)-f_{1,0}(s)\textup{d}s\big|+\big|\int_{t}^{1} f_{0,1}(s)-f_{1,1}(s)\textup{d}s\big|\Big)\\
&\leq\frac{1}{2}\Big(\int_{t}^{1} \big|f_{0,0}(s)-f_{1,0}(s)\big|\textup{d}s+\int_{t}^{1} \big|f_{0,1}(s)-f_{1,1}(s)\big|\textup{d}s\Big)\\
&\leq\frac{1}{2}(\epsilon_0+\epsilon_1)(1-t).
\end{aligned}
\label{eq:eo1}
\end{equation}
where we define the upper bounds $\big|f_{0,0}(s)-f_{1,0}(s)\big |\leq\epsilon_0$ and $\big|f_{0,1}(s)-f_{1,1}(s)\big |\leq\epsilon_1, \forall s\in[0,1]$. Also, we have
\begin{equation}
\begin{aligned}
\Delta \textup{EO}&=\frac{1}{2}\Big(\big |P_{0,0}(\hat{Y}=1)-P_{1,0}(\hat{Y}=1)\big |+\big |P_{0,1}(\hat{Y}=1)-P_{1,1}(\hat{Y}=1)\big|\Big)\\
&=\frac{1}{2}\Big(\big |\big(1-P_{0,0}(\hat{Y}=0)\big)-\big(1-P_{1,0}(\hat{Y}=0)\big)\big |+\big |\big(1-P_{0,1}(\hat{Y}=0)\big)-\big(1-P_{1,1}(\hat{Y}=0)\big)\big|\Big)\\
&=\frac{1}{2}\Big(\big |P_{1,0}(\hat{Y}=0)-P_{0,0}(\hat{Y}=0)\big |+\big |P_{1,1}(\hat{Y}=0)-P_{0,1}(\hat{Y}=0)\big|\Big)\\
&=\frac{1}{2}\Big(\big|\int_{0}^{t} f_{1,0}(s)-f_{0,0}(s)\textup{d}s\big|+\big|\int_{0}^{t} f_{1,1}(s)-f_{0,1}(s)\textup{d}s\big|\Big)\\
&\leq\frac{1}{2}\Big(\int_{0}^{t} \big|f_{1,0}(s)-f_{0,0}(s)\big|\textup{d}s+\int_{0}^{t} \big|f_{1,1}(s)-f_{0,1}(s)\big|\textup{d}s\Big)\\
&\leq\frac{1}{2}(\epsilon_0+\epsilon_1)t.
\end{aligned}
\label{eq:eo2}
\end{equation}

Combining (\ref{eq:eo1}) and (\ref{eq:eo2}), we have
\begin{equation}
\Delta \textup{EO}\leq\frac{1}{2}(\epsilon_0+\epsilon_1)\cdot\min\{t,1-t\}\leq\frac{1}{4}(\epsilon_0+\epsilon_1).
\end{equation}

\end{appendix}

\end{document}